\documentclass[letterpaper]{article} %
\usepackage{aaai25}  %
\usepackage{times}  %
\usepackage{helvet}  %
\usepackage{courier}  %
\usepackage[hyphens]{url}  %
\usepackage{graphicx} %
\usepackage{natbib}  %
\usepackage{caption} %
\usepackage{algorithm}
\usepackage{algorithmic}

\usepackage{xspace}
\usepackage{xcolor}
\usepackage{subcaption}
\usepackage{amssymb}
\usepackage{amsmath}
\usepackage[capitalise]{cleveref} %
\crefname{figure}{Fig.}{Figs.}
\Crefname{figure}{Fig.}{Figs.}
\crefname{table}{Tab.}{Tabs.}
\Crefname{table}{Tab.}{Tabs.}
\Crefname{section}{\S}{\S}
\usepackage{booktabs}
\usepackage{array, makecell} %
\usepackage{multirow}
\definecolor{olivegreen}{rgb}{0, 0.6, 0}
\definecolor{black}{HTML}{000000}
\definecolor{white}{HTML}{ffffff}
\definecolor{color1}{HTML}{ACE5EE}
\definecolor{color2}{HTML}{0093AF}
\definecolor{color3}{HTML}{CC0000}%
\definecolor{color4}{HTML}{0087BD}
\definecolor{color5}{HTML}{333399}
\definecolor{color6}{HTML}{20B2AA}

\usepackage{nicefrac}       %

\usepackage{pifont}%
\newcommand{\cmark}{\color{olivegreen}\ding{51}}%
\newcommand{\xmark}{\color{red}\ding{55}}%

\newcommand{\aname}{MimiQ\xspace}
\newcommand{\JL}[1]{{\color{olivegreen}[\textbf{\sc JLee}: \textit{#1}]}}
\newcommand{\KH}[1]{{\color{purple}[\textbf{\sc KH}: \textit{#1}]}}
\newcommand{\HY}[1]{{\color{blue}[\textbf{\sc HY}: \textit{#1}]}}
\newcommand{\NS}[1]{{\color{red}[\textbf{\sc Noseong}: \textit{#1}]}}
\newcommand{\SJ}[1]{{\color{orange}[\textbf{\sc SJ}: \textit{#1}]}}
\newcommand{\DI}[1]{{\color{aquamarine}[\textbf{\sc DI}: \textit{#1}]}}

\def\final{}   %
\ifdefined\final
\renewcommand{\JL}[1]{}
\renewcommand{\KH}[1]{}
\renewcommand{\HY}[1]{}
\renewcommand{\NS}[1]{}
\renewcommand{\SJ}[1]{}
\renewcommand{\DI}[1]{}

\fi

\usepackage{cuted}

\newcommand{\loss}{\mathcal{L}}
\newcommand{\round}[1]{\ensuremath{\lfloor#1\rceil}}

\usepackage{newfloat}
\usepackage{listings}
\DeclareCaptionStyle{ruled}{labelfont=normalfont,labelsep=colon,strut=off} %
\floatstyle{ruled}
\newfloat{listing}{tb}{lst}{}
\floatname{listing}{Listing}
\title{\aname: Low-Bit Data-Free Quantization of Vision Transformers\\with Encouraging Inter-Head Attention Similarity\\(Extended Version)}
\author {
    Kanghyun Choi\textsuperscript{\rm 1},
    Hyeyoon Lee\textsuperscript{\rm 1},
    Dain Kwon\textsuperscript{\rm 1},
    SunJong Park\textsuperscript{\rm 1},
    Kyuyeun Kim\textsuperscript{\rm 2},
    Noseong Park\textsuperscript{\rm 3},
    Jonghyun Choi\textsuperscript{\rm 1},
    Jinho Lee\textsuperscript{\rm 1}\thanks{Corresponding author}
}
\affiliations {
    \textsuperscript{\rm 1}Seoul National University \\
    \textsuperscript{\rm 2}Google \\
    \textsuperscript{\rm 3}KAIST \\
    \{kanghyun.choi, hylee817, dain.kwon, ryan0507, jonghyunchoi, leejinho\}@snu.ac.kr, \\ kyuyeunk@google.com, noseong@kaist.ac.kr
}

\usepackage{bibentry}

\begin{document}

\maketitle

\begin{abstract}
Data-free quantization (DFQ) is a technique that creates a lightweight network from its full-precision counterpart without the original training data, often through a synthetic dataset.
Although several 
DFQ methods have been proposed for vision transformer (ViT) architectures, they fail to achieve efficacy in low-bit settings.
Examining the existing methods, we observe that their synthetic data produce misaligned attention maps,
while those of the real samples are highly aligned.
From this observation, we find that aligning attention maps of synthetic data helps improve the overall performance of quantized ViTs.
Motivated by this finding, we devise \aname, a novel DFQ method designed for ViTs that
enhances
inter-head attention similarity. 
First, we generate synthetic data by aligning head-wise attention outputs from each spatial query patch. 
Then, we align the attention maps of the quantized network to those of the full-precision teacher by applying head-wise structural attention distillation. 
The experimental results show that the proposed method significantly outperforms baselines, setting a new state-of-the-art for ViT-DFQ. %
This paper is an extended version of our work published in the proceedings of AAAI 2025, including additional supplementary material.
\end{abstract}

\begin{links}
    \link{Code}{https://github.com/iamkanghyunchoi/mimiq}
\end{links}

\section{Introduction}
Over the past few years, Vision Transformers (ViT)~\cite{vit} have gained increasing interest due to their remarkable performance on many computer vision tasks.
However, ViT has high computational costs compared to conventional CNNs, making it challenging to adopt in many resource-constrained devices. %
Thus, various works focus on reducing the costs of ViT architectures~\cite{qvit, ptqvit, spvit, univit}.
One popular approach is network quantization~\cite{nagel2021white, gholami2021survey}, which converts floating-point parameters and features to low-bit integers.
However, naively converting the parameters to lower-bit induces a large accuracy drop, which is why quantization usually requires additional calibration~\cite{sung2015resiliency,ptqvit,fqvit} or fine-tuning~\cite{binaryconnect, hubara2017quantized} using the original training dataset. 
Unfortunately, in real-life cases, 
the original training dataset is not always available
due to privacy concerns, security issues, or copyright protections~\cite{liu2021machine, hathaliya2020exhaustive}.

\begin{figure}[t]
\centering
\includegraphics[width=.85\columnwidth]{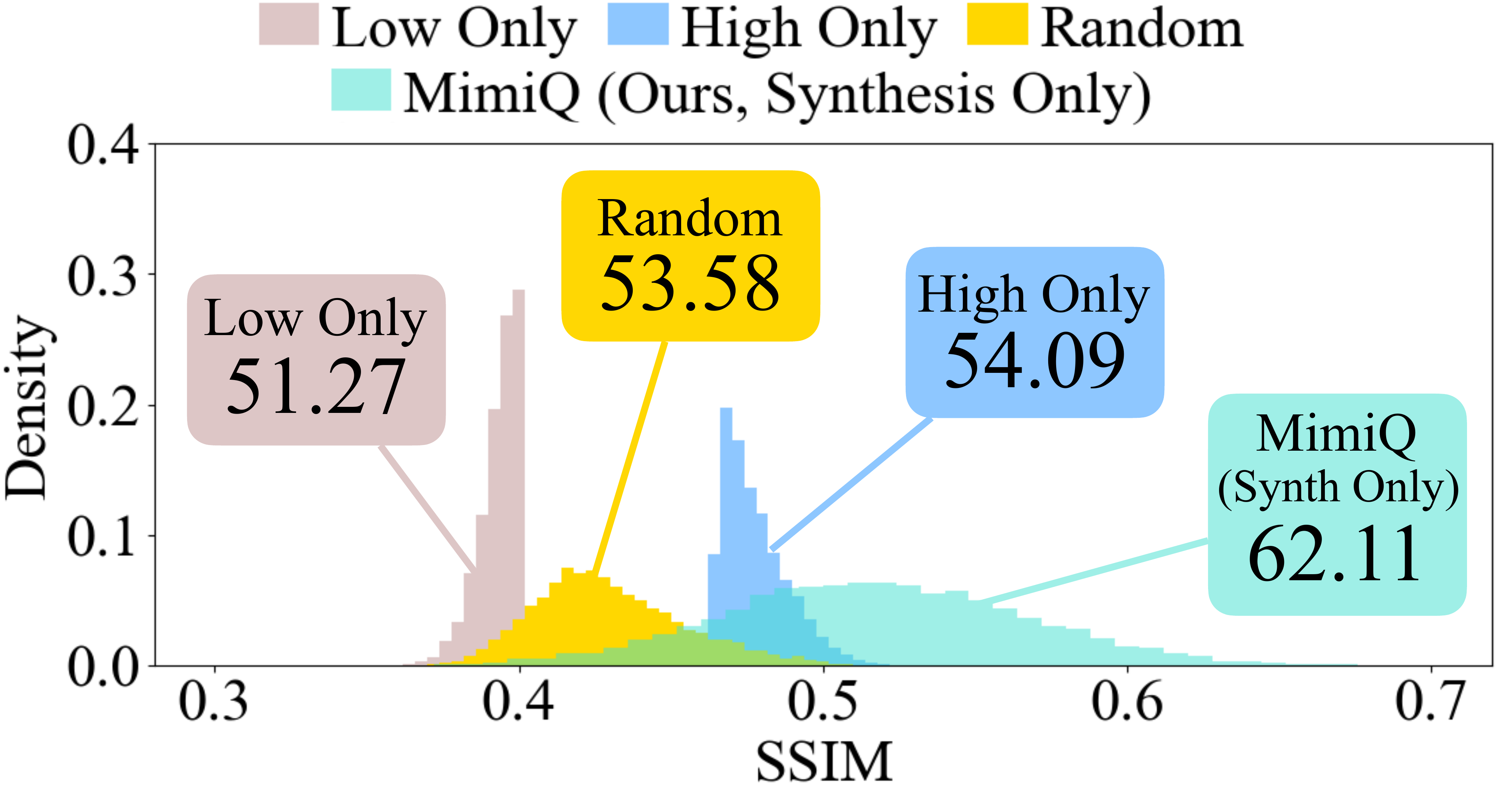}
\caption{Attention similarity histograms of Base DF Synthesis (high and low similarity, random sampled) and \aname.
Colored boxes denote ImageNet accuracy of the corresponding dataset.
This motivational study shows the attention similarity is related to the DFQ accuracy of ViTs.
}
\label{fig:moti:hist}
\end{figure}

Data-free quantization (DFQ)~\cite{dfq} addresses such dataset inaccessibility by 
quantizing the network without using the original training data. 
To replace the original dataset, 
recent methods generate synthetic data from the
pretrained networks
and use it for calibration.
These approaches directly optimize synthetic samples with gradient descent~\cite{intraq,psaq,psaqv2} or train an auxiliary data generator~\cite{gdfq, qimera}. %

Unfortunately, existing
 DFQ methods suffer from destructive accuracy drops on low-bit ViTs (refer to \cref{tab:master}). %
This is because CNN-based DFQs rely on batch normalization (BN) statistics to create synthetic samples resembling the original training data, making them unsuitable for ViTs that do not contain the BN layer. %
Recent DFQ approaches for ViTs, such as PSAQ V1/V2~\cite{psaq,psaqv2}, employ a patch similarity metric to separate foreground from background in images. 
However, these methods overlook the overall image structure and the positional context of patches, potentially resulting in poor quality synthetic images. 

To this end, we propose \aname, a DFQ framework for low-bit ViT quantization by focusing on inter-head attention similarity. %
By inspecting the attention maps of real and synthetic data, we observe that synthetic samples show misaligned attention maps, and aligning those maps improves accuracy (\cref{fig:moti:hist}). %
Inspired by this motivational study, we design a DFQ method to achieve inter-head attention similarity. %
In data generation, we align inter-head attention maps of synthetic samples by minimizing the distance between head-wise maps from each spatial query patch. 
For fine-tuning, we employ head-wise structural attention distillation on the quantized network to mimic its full-precision counterpart. 

We extensively evaluate \aname across various tasks, ViT networks, and bit settings. 
The experimental results show that \aname outperforms baselines by a significant margin especially in low-bit settings, 
reducing the gap between data-free and real-data quantization. 
As a result, \aname sets a new state-of-the-art for the data-free ViT quantization. 

Our primary contributions are summarized as follows:
\begin{itemize} %
\item We discover DFQ baselines produce misaligned attention maps across attention heads, and aligning attention maps contributes to the quantization accuracy.

\item We propose a synthetic data generation method to align inter-head attention by reducing the structural distance between attention heads output from each query patch. %

\item We propose a head-wise attention distillation method aligning the structure of attention outputs of quantized networks with those of full-precision teachers. %

\item 
The experiments on various tasks and ViT architectures show that \aname achieves new state-of-the-art performance for data-free ViT quantization.
\end{itemize}

\section{Backgrounds}
\label{sec:backgrounds}

\subsection{ViT Architectures and Multi-Head Attention}
\label{sec:vit}

ViT~\cite{vit} is an adaptation of Transformer from NLP~\cite{vaswani2017attention} to vision.
Each Transformer block comprises 
a multi-head self-attention (MSA) layer and a feed-forward layer. 
For the length $N_d$ input sequence with $d$-dimension, $X_{\in N_d\times d}$, MSA performs attention using multiple heads to obtain diverse features as:
\begin{align}
    MSA(X) &= [H_1(X),\cdots,H_N(X)] W^O,
    \label{eq:mhsa1} 
\end{align}
where $N$ is the number of attention heads. 
The outputs of each head are concatenated ($[\cdot]$) and merged by multiplication with projection matrix $W^O$.
Each attention head has separated weights ($W^Q_h, W^K_h, W^V_h$) for computing query, key, and value vectors. 
The output of $h$-th head is as follows:
\begin{align}
    (Q_h,K_h,V_h) &= (XW^Q_h, XW^K_h, XW^V_h) \label{eq:qkv} \\
    H_h(X) &= softmax(\frac{Q_h K_h^\intercal}{\sqrt{d}})V_h.
    \label{eq:mhsa2}
\end{align}

\subsection{Data-Free Quantization}
\label{sec:dfq}

\textbf{Quantization} 
reduces network complexity by converting floating-point to integer operations
~\cite{nagel2021white,gholami2021survey}.
We employ uniform quantization
which uses a simple scale ($s$) and zero-point ($z$) mapping to transform floating-point values $\theta$ into integers $\theta^{int}$:
\begin{align}
    \theta^{int} &= clamp(\round{\theta \cdot s - z}, q_{min}, q_{max}), 
    \label{eq:q_basic}
\end{align}
where 
($q_{min}$,~$q_{max}$) indicates minimum and maximum of the integer representation range, i.e., ($-2^{k-1}$,~$2^{k-1}-1$).
Refer to the supplementary material for more details and note that this work is not limited to a certain quantization scheme.
One drawback of quantization is the potential accuracy loss due to reduced precision. 
To counter this, quantization-aware training (QAT) uses fine-tuning to regain lost accuracy.  %
However, the training dataset is not always available in real-world scenarios~\cite{liu2021machine, hathaliya2020exhaustive}, making QAT inapplicable. 

\begin{figure*}[t]
\centering
\includegraphics[width=.82\textwidth]{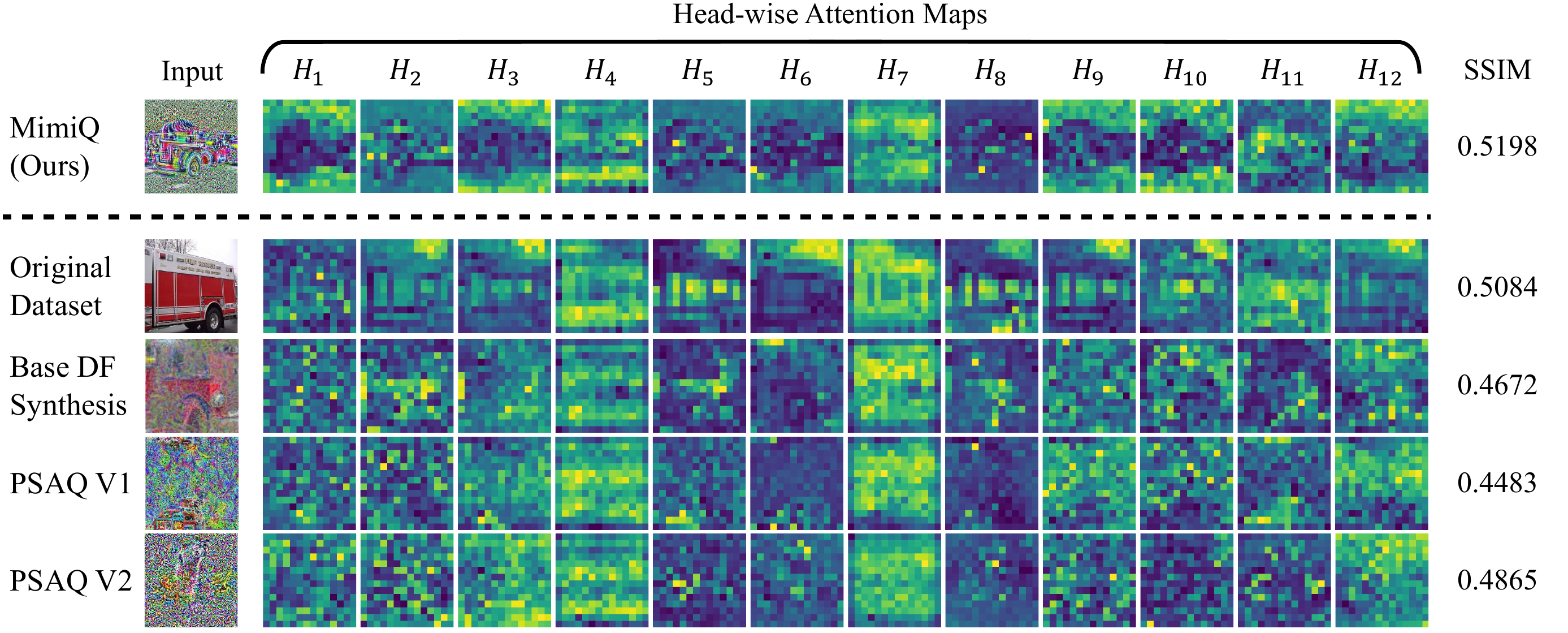}
\caption{Attention visualization of synthetic samples and the original dataset.
Compared to the original dataset, synthetic data from baselines present misaligned attention maps. 
The first row, a sample generated by \aname, shows aligned attention maps across attention heads.
The measured average attention similarity (SSIM) further validates \aname shows the best alignment.
}
\label{fig:motiv_map}
\end{figure*}

\textbf{Data-Free Quantization} aims to quantize pretrained networks without access to any of the real training data,
mostly by using synthetic samples as surrogates.
The major challenge is that one cannot use the training data or external generators for the synthesis, as it would fall into a case of data leakage.
Instead, information from pretrained full-precision networks $f$ is used by optimizing the following terms:
\begin{align}
    \mathcal{L}_{CL} &= - \textstyle \sum\nolimits_{c=1}^{C} \hat y_c log(f(I)_c), \label{eq:aux_cl} \\
    \mathcal{L}_{TV} &= \lvert\lvert I_{h,w} - I_{h+1,w}  \lvert\lvert^2_2 + \lvert\lvert I_{h,w} - I_{h,w+1}  \lvert\lvert^2_2, \label{eq:aux_tv} \\
    \mathcal{L}_{BNS} &= \lVert \hat \mu_{l} - \mu_{l} \lVert^2_2 + \lVert \hat \sigma_{l} - \sigma_{l} \lVert^2_2, \label{eq:aux_bns}
\end{align}
where $I$ is the synthetic image, $\hat y_c$ is a class label among $C$ classes, ($h$,$w$) are pixel coordinates, and ($\mu_l$,$\sigma_l$) are BN statistics of the $l$-th layer. 
$\mathcal{L}_{CL}$ embeds prior knowledge of the pretrained classifier, and $\mathcal{L}_{TV}$ prevents steep changes between nearby pixels. 
$\mathcal{L}_{BNS}$ reduces the distance between feature statistics of synthetic samples and BN layers, %
but it is not applicable to ViTs due to its lack of BN layers. %

\section{Related Work}
\label{sec:related}

\subsection{Vision Transformer Quantization}
After the success of ViTs, many efforts have been followed to reduce its computational and memory costs through quantization.
One of the pioneering efforts is PTQ-ViT~\cite{ptqvit}, which 
performed quantization to preserve the functionality of the attention.
Then followed FQ-ViT~\cite{fqvit} proposed to fully quantize ViT, including LayerNorm and Softmax.
PTQ4ViT~\cite{yuan2022ptq4vit} applied twin uniform quantization strategy and Hessian-based metric for determining scaling factor. 
I-ViT~\cite{ivit} performed integer-only quantization without any floating-point arithmetic. %
Q-ViT~\cite{qvit} and RepQ-ViT~\cite{repqvit} 
proposed remedies to overcome accuracy degradation in low-bit ViTs.
However, they require the original training data for calibration, and do not consider real-world scenarios where training data is often unavailable.

\subsection{Data-Free Vision Transformer Quantization}
After the first proposal for data-free quantization~\cite{dfq},
many efforts specialized for CNNs have followed, including
ZeroQ~\cite{zeroq}, DSG~\cite{dsg}, and intraQ~\cite{intraq}.
Notably, GDFQ~\cite{gdfq} proposed to jointly train generators to synthesize samples, which
laid the foundation for variants using better generators~\cite{autorecon}, boundary samples~\cite{qimera}, smooth loss surface~\cite{ait}, and sample adaptability~\cite{adadfq}.
This stream of methods owes a large portion of its success to the BN statistics (\cref{eq:aux_bns}).

Unfortunately, the BN layer is absent in ViTs, making the CNN-targeted techniques
suffer from inaccurate sample distribution when adopted to ViTs.
DFQ for ViT is 
still in an early stage of development. 
PSAQ-ViT~\cite{psaq} was the first to present DFQ for ViTs, utilizing inter-patch discrepancy of foreground and background patches to generate realistic samples. 
The following work, PSAQ-ViT V2~\cite{psaqv2}, uses an adaptive method by applying the concept of adversarial training.
However, they only focus on patch-level similarity, neglecting overall image structure.
Parallel to our work, \citet{ramachandran2024clamp}, recently disclosed online, explores patch-level contrastive learning to enhance the synthetic data, but still only considers patch-level information similar to prior works.
As their method does not consider low-bit regions and has different quantization settings from ours, we believe it is not quite adequate to compare ours with their reported results.

\begin{figure*}[t]
\centering
\includegraphics[width=0.8\textwidth]{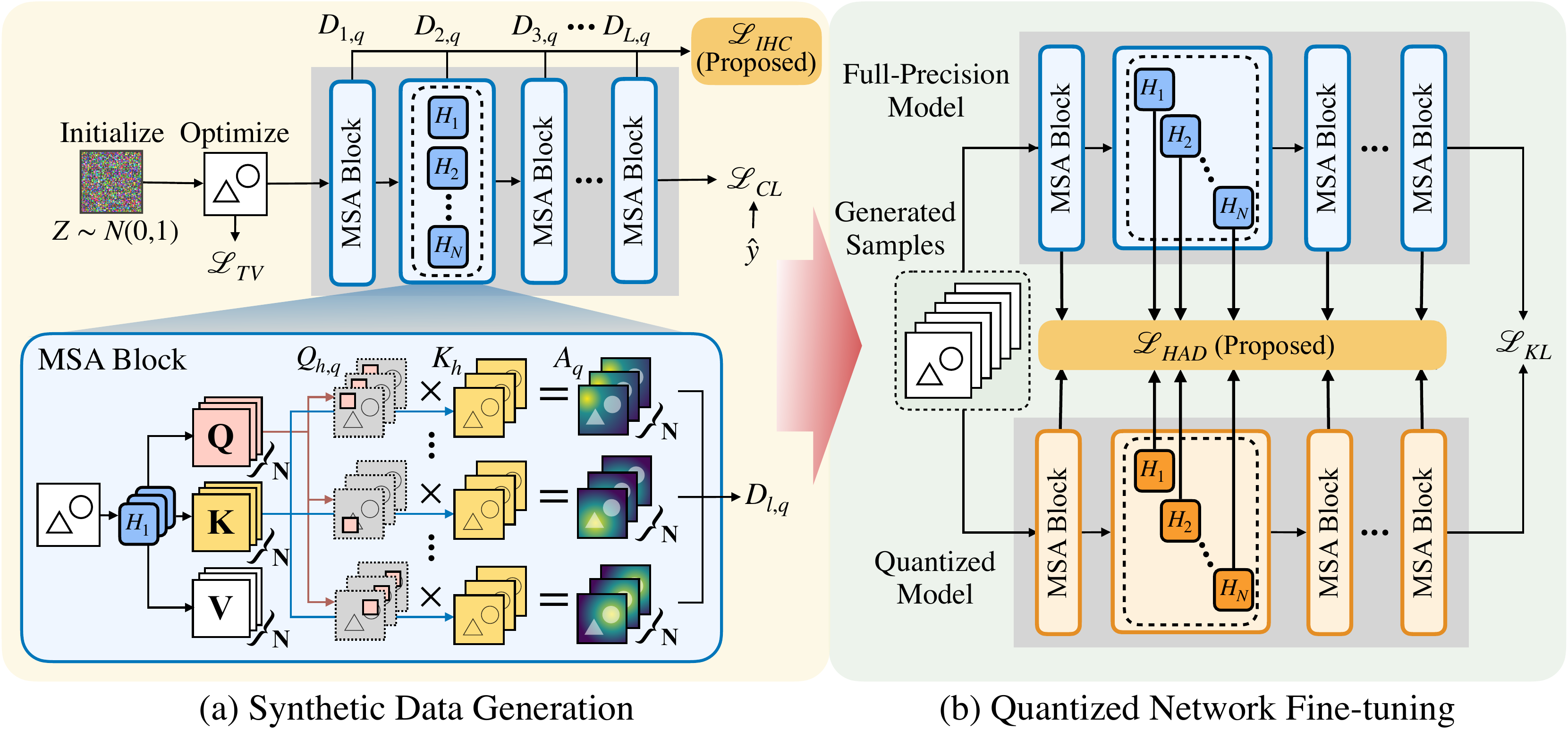}
\caption{An overview of the proposed method. 
(a) The synthetic samples are initialized with Gaussian noise, then optimized with $\mathcal{L}_G$ (\cref{eq:lg}). 
The proposed inter-head coherency loss $\mathcal{L}_{IHC}$ measures the similarity of head-wise attention maps from the same query patch index. 
(b) The quantized network is trained to minimize output ($\mathcal{L}_{KL}$) and inter-head ($\mathcal{L}_{HAD}$) discrepancy.
}
\label{fig:overview}
\end{figure*}

\section{Motivational Study}
\label{sec:motiv}

All the baseline DFQ methods experience huge accuracy drops in low-bit settings compared to the real-data QAT (\cref{tab:master}). %
To investigate the source of such discrepancy, 
we 
inspect attention maps of ViTs, %
specifically
the ViT's head-wise attention maps, denoted as $H_i$ in \cref{eq:mhsa1}.

The visualization in \cref{fig:motiv_map} shows that real samples lead to similarly structured attention maps, unlike data-free synthetic samples.
We set the base DF synthesis method, which generates synthetic samples with $\mathcal{L}_{CL}$ (\cref{eq:aux_cl}) and $\mathcal{L}_{TV}$ (\cref{eq:aux_tv}),
where we also analyze PSAQ V1 and V2 as additional baselines. %
On the one hand, the attention maps from real images clearly display the object's structure in most heads.
While minor variations exist in different heads highlighting different parts, either the object itself (e.g., $H_{11}$) or the background (e.g., $H_9$), they exhibit high similarity to one another.
On the other hand, synthetic samples from baselines do not seem to produce visually similar features across attention heads. %
In addition, we present a quantitative analysis with the SSIM score. 
In the last column of \cref{fig:motiv_map}, the baselines show lower attention similarity compared to the real samples.
The results also show that \aname show the best alignment of attention heads. 
\Cref{fig:motiv_map} shows visualizations from ViT-Base architecture with 12 attention heads.
Please refer to the supplementary material for more results. %

From the observation in \cref{fig:motiv_map}, we hypothesize that aligning inter-head attention from synthetic samples contributes to better accuracy of data-free quantized ViTs. 
To validate this, we performed motivational experiments to identify the correlation between attention map similarity and quantization accuracy. 
First, we generate a synthetic dataset using the base DF synthesis mentioned above. 
We then measure the inter-head attention similarity of each image with structural similarity index measure (SSIM, \citet{ssim}) and construct subsets of the synthetic dataset having 1) high attention similarity and 2) low attention similarity. 
For comparison, we construct a control group with 3) random sampling.
Lastly, we train a W4/A4 quantized ViT network with each sampled group and examine the accuracy.

The results of the experiments (\cref{fig:moti:hist}) show that the quantized networks trained with samples of low attention similarity consistently underperform compared to networks trained with samples of high attention similarity. 
These results empirically validate our hypothesis that the inter-head attention similarity of synthetic samples correlates with the quantization accuracy.
Based on the observation, we devise \aname to encourage inter-head attention similarity throughout the whole DFQ process, including both data generation and fine-tuning phases.
As shown with turquoise bars in \cref{fig:moti:hist}, samples generated with \aname yield higher attention similarity and have superior quantization accuracy.

\section{Proposed Method}
\label{sec:method}

Inspired by the observation from the motivational study, we propose \aname framework, a DFQ framework for ViTs that utilizes head-wise similarity information from MSA layers. 
The overall process is depicted in \cref{fig:overview}.
We promote inter-head similarity in two directions of sample synthesis and head-wise distillation during fine-tuning.

\subsection{Sample Synthesis with Inter-Head Similarity}
\label{sec:img_gen}

We propose synthetic sample generation with inter-head attention similarity by aligning head-wise internal attention maps. %
The \cref{fig:overview}a depicts attention head alignment.
First, we collect attention maps from each head that shares the same spatial query patch index $q$ out of $P$ patches:
\begin{align}
    A_{q} = 
    \begin{bmatrix}
    Q_{1,q}K_1^\intercal & Q_{2,q}K_2^\intercal & \cdots & Q_{N,q}K_N^\intercal
\end{bmatrix},
    \label{eq:img_synth_coll}
\end{align}
where $A_q$ is the collected attention map and ($Q$,$K$) are query and key matrices from \cref{eq:qkv}, respectively. 
Then, we measure the average distance $D_q$ with attention distance metric $f_{dist}$ between the attention heads as follows: 
\begin{align}
     D_q = \frac{1}{N^2} \textstyle\sum^N_i \sum^N_j f_{dist}(A_{q,i}, A_{q,j}).
    \label{eq:img_synth_avg}
\end{align}

Here, $f_{dist}$ needs to consider the nature of attention maps:
Attention maps can be inverted while retaining the structure of images. 
For example, in the original sample in \cref{fig:motiv_map},
$H_{11}$ focuses on an object while $H_9$ focuses on the background. 
The metric should consider such relation among the heads.

To this end, we use absolute SSIM as $f_{dist}$.
A higher magnitude of SSIM indicates a higher correlation in both the positive and negative (inverted) directions, representing the structural similarity between two attention maps.
Using \cref{eq:img_synth_avg}, we can optimize synthetic samples towards inter-head attention similarity with synthesis loss $\mathcal{L}_{G}$. 
\begin{align}
    \mathcal{L}_{G} = \mathcal{L}_{IHC} + \alpha \mathcal{L}_{CL} + \beta \mathcal{L}_{TV}, \label{eq:lg}\\
    \mathcal{L}_{IHC} = \frac{1}{LP} \textstyle\sum^L_l \sum^{P}_q (1-D_{l,q}),
\end{align}
where $\alpha$ and $\beta$ are hyperparameters for $\mathcal{L}_{CL}$ (\cref{eq:aux_cl}) and $\mathcal{L}_{TV}$ (\cref{eq:aux_tv}), respectively, $L$ is the number of MSA layers in the model, %
and $D_{l,q}$ is $D_q$ from layer $l$.
The generated synthetic samples can be found in \cref{fig:gradcam_lpips} and Appendix in the supplementary material. 

\begin{table}[]
    \centering

     \resizebox{0.85\columnwidth}{!}
     {
    \begin{tabular}{lcccc}
    \toprule
    {Corr. Coeff.} & DSSIM & MSE & L1-Dist. & KL-Div. \\
    \midrule

Spearman	&	\textbf{0.9970} & 0.9823 & 0.9796 & 0.9657	\\
																		
Kendall	&	\textbf{0.9520} & 0.8937 & 0.8863 & 0.8566	\\

      \bottomrule
    \end{tabular} 
     } %
    \caption{Absolute correlation coefficients of attention coherency metric to the quantized accuracy.}
    \label{tab:att_dist_corr}
\end{table}

\begin{table*}[t]
    \centering

    \resizebox{.8\textwidth}{!}
    {
    \begin{tabular}{clcccccccccc}
    \toprule
    \multirow{2}{*}{Bits} & \multirow{2}{*}{Methods} & \multirow{2}{*}{\makecell{Target \\ Arch.}} &  \multicolumn{9}{c}{Networks} \\
    \cmidrule{4-12}  
       & &  &  ViT-T & ViT-S  & ViT-B & DeiT-T & DeiT-S  & DeiT-B & Swin-T & Swin-S  & Swin-B \\
    \midrule
\multirow{8}{*}{W4/A4}& Real-Data FT & - &		58.17	&	67.21	&	67.81	&	57.98	&	62.15	&	64.96	&	73.08	&	76.34	&	73.06	\\ \cmidrule{2-12}						
& GDFQ & CNN&		\textcolor{white}02.95	&	\textcolor{white}04.62	&	11.73	&	25.96	&	22.12	&	30.04	&	42.08	&	41.93	&	36.04	\\ 						
& Qimera & CNN&		\textcolor{white}00.57	&	\textcolor{white}07.02	&	\textcolor{white}05.61	&	15.18	&	11.37	&	32.49	&	47.98	&	39.64	&	29.27	\\ 						
& AdaDFQ & CNN&		\textcolor{white}02.00	&	\textcolor{white}01.78	&	\textcolor{white}06.21	&	19.57	&	14.44	&	19.22	&	38.88	&	39.40	&	32.26	\\						
&PSAQ-ViT V1 & ViT &		\textcolor{white}00.67	&	\textcolor{white}00.15	&	\textcolor{white}00.94	&	19.61	&	\textcolor{white}05.90	&	\textcolor{white}08.74	&	22.71	&	\textcolor{white}09.26	&	23.69	\\ 						
& PSAQ-ViT V2 & ViT &		\textcolor{white}01.54	&	\textcolor{white}04.14	&	\textcolor{white}02.83	&	22.82	&	32.57	&	45.81	&	50.42	&	39.10	&	39.26	\\ \cmidrule{2-12}						
&{\aname (Ours)}  & \multirow{2}{*}{ViT} &		\textbf{42.99}	&	\textbf{55.69}	&	\textbf{62.91}	&	\textbf{52.03}	&	\textbf{62.72}	&	\textbf{74.10}	&	\textbf{69.33}	&	\textbf{70.46}	&	\textbf{73.49}	\\						
&Acc. Gain&&		+40.04	&	+48.68	&	+51.18	&	+26.07	&	+30.15	&	+28.28	&	+18.91	&	+28.53	&	+34.23	\\ \midrule						
																									
\multirow{8}{*}{W5/A5}& Real-Data FT & - &		68.49	&	73.90	&	80.52	&	66.10	&	73.95	&	78.39	&	78.71	&	81.74	&	83.08	\\ \cmidrule{2-12}						
& GDFQ & CNN&		24.40	&	53.96	&	33.56	&	44.76	&	57.00	&	71.03	&	61.30	&	78.04	&	70.55	\\ 						
& Qimera & CNN&		26.70	&	16.13	&	\textcolor{white}09.43	&	33.13	&	33.65	&	47.01	&	62.13	&	46.81	&	43.57	\\ 						
& AdaDFQ & CNN&		27.10	&	59.36	&	43.02	&	53.85	&	59.55	&	71.12	&	64.61	&	79.82	&	75.59	\\ 						
&PSAQ-ViT V1 & ViT &		17.66	&	23.37	&	16.80	&	53.36	&	47.35	&	57.23	&	58.63	&	76.33	&	57.80	\\ 						
& PSAQ-ViT V2 & ViT &		40.21	&	63.59	&	74.29	&	55.18	&	65.30	&	73.16	&	69.77	&	80.55	&	79.80	\\ \cmidrule{2-12}						
&{\aname (Ours)}  & \multirow{2}{*}{ViT} &		\textbf{62.40}	&	\textbf{70.02}	&	\textbf{78.09}	&	\textbf{63.40}	&	\textbf{72.59}	&	\textbf{78.20}	&	\textbf{76.39}	&	\textbf{80.75}	&	\textbf{82.05}	\\						
&Acc. Gain&&		+22.19	&	+6.43	&	+3.80	&	+8.22	&	+7.28	&	+5.04	&	+6.63	&	+0.20	&	+2.25	\\ \midrule						
																									
\multirow{8}{*}{W4/A8}& Real-Data FT & - &		71.52	&	79.84	&	84.52	&	70.37	&	78.93	&	81.47	&	80.47	&	82.46	&	84.29	\\ \cmidrule{2-12}						
& GDFQ & CNN&		62.65	&	76.06	&	81.68	&	65.82	&	76.49	&	80.03	&	78.90	&	81.47	&	83.63	\\ 						
& Qimera & CNN&		61.80	&	60.08	&	63.22	&	61.90	&	70.10	&	72.38	&	73.93	&	72.22	&	76.35	\\ 						
& AdaDFQ & CNN&		64.67	&	76.27	&	82.43	&	67.71	&	76.92	&	80.49	&	79.70	&	82.07	&	83.78	\\ 						
&PSAQ-ViT V1 & ViT &		59.59	&	62.98	&	67.74	&	66.16	&	76.56	&	80.05	&	79.06	&	81.89	&	79.51	\\ 						
& PSAQ-ViT V2 & ViT &		66.78	&	78.24	&	84.02	&	68.23	&	78.27	&	81.15	&	79.98	&	82.04	&	{83.90}	\\ \cmidrule{2-12}						
&{\aname (Ours)}  & \multirow{2}{*}{ViT} &		\textbf{68.15}	&	\textbf{78.77}	&	\textbf{84.20}	&	\textbf{69.86}	&	\textbf{78.48}	&	\textbf{81.34}	&	\textbf{80.06}	&	\textbf{82.08}	&	\textbf{83.99}	\\						
&Acc. Gain&&		+1.37	&	+0.53	&	+0.18	&	+1.63	&	+0.21	&	+0.20	&	+0.08	&	+0.01	&	+0.09	\\ \midrule						
																									
\multirow{8}{*}{W8/A8}& Real-Data FT & - &		74.83	&	81.30	&	85.13	&	71.99	&	79.70	&	81.77	&	80.96	&	83.08	&	84.79	\\ \cmidrule{2-12}						
& GDFQ & CNN&		72.90	&	80.97	&	84.81	&	71.83	&	79.59	&	81.62	&	80.83	&	82.99	&	84.42	\\ 						
& Qimera & CNN&		72.88	&	81.04	&	84.98	&	71.76	&	79.46	&	81.58	&	80.41	&	82.95	&	84.37	\\ 						
& AdaDFQ & CNN&		73.84	&	81.11	&	84.88	&	71.72	&	79.34	&	81.73	&	80.89	&	82.99	&	84.70	\\ 						
&PSAQ-ViT V1 & ViT &		72.73	&	81.17	&	84.89	&	{71.99}	&	{79.71}	&	81.79	&	\textbf{81.26}	&	\textbf{83.29}	&	\textbf{85.13}	\\ 						
& PSAQ-ViT V2 & ViT &		73.43	&	81.25	&	85.11	&	71.90	&	79.70	&	81.86	&	80.88	&	83.00	&	84.71	\\ \cmidrule{2-12}						
&{\aname (Ours)}  & \multirow{2}{*}{ViT} &		\textbf{74.60}	&	\textbf{81.30}	&	\textbf{85.17}	&	\textbf{72.01}	&	\textbf{79.73}	&	\textbf{81.87}	&	80.96	&	83.07	&	84.79	\\						
&Acc. Gain&&		+0.76	&	+0.05	&	+0.07	&	+0.02	&	+0.02	&	+0.01	&	-0.30	&	-0.22	&	-0.34	\\ \bottomrule						

    \end{tabular}}
   
        \caption{Comparison on  ImageNet image classification dataset.}

    \label{tab:master}
\end{table*}

\subsection{Head-Wise Structural Attention Distillation}
\label{sec:ssim_dist}

Here, we propose head-wise structural distillation from a full-precision teacher shown in \cref{fig:overview}b, in addition to utilizing our attention-aligned samples.
Along with the output matching loss (i.e., $\mathcal{L}_{KL}$) commonly adopted for QAT, we further reduce the distance $g_{dist}(\cdot)$ between each attention output pair by optimizing the following objective:
\begin{align}
    \mathcal{L}_{HAD} = \frac{1}{LN} \textstyle\sum^L_l \sum^N_i g_{dist}(H^\mathcal{T}_{l,i}, H^\mathcal{S}_{l,i}),
    \label{eq:attn_distill_avg}
\end{align}
where $\loss_{HAD}$ is head-wise attention distillation loss.
$H^\mathcal{T}_{l, i}$ and $H^\mathcal{S}_{l, i}$ are $i$-th attention head outputs from the $l$-th layer of teacher and student, respectively. 
Therefore, the training objective $\loss_T$ of the quantized network is as follows:
\begin{align}
    \mathcal{L}_{T} &= \mathcal{L}_{KL}(f_\mathcal{T}(\hat{X}) \lvert \lvert f_\mathcal{S}(\hat{X})) + \gamma \mathcal{L}_{HAD},
    \label{eq:pch2hd}
\end{align}
where $\hat{X}$ is synthetic samples, $\gamma$ is a hyperparameter. %

We compare four candidate metrics for $g_{dist}$: Mean-squared error (MSE), L1 distance, KL-divergence, and structural dissimilarity (DSSIM), i.e., the negative of SSIM. %
To choose a metric relevant to quantization accuracy, we randomly quantized a portion of attention heads in each MSA layer of the pretrained ViT-Base network and measure the attention head distance and network accuracy.
We sample 500 settings from the configuration space and report Spearman and Kendall rank correlation coefficients.
The comparison is shown in \cref{tab:att_dist_corr}.
The results show that DSSIM has the highest correlation with quantized accuracy. 
According to the experimental results, we choose DSSIM as $g_{dist}$.
Please refer to supplementary material for data visualization.

\section{Performance Evaluation}
\label{sec:experiments}

\subsection{Experimental Settings}

We evaluate \aname using tiny, small, and base versions of %
ViT~\cite{vit}, DeiT~\cite{deit}, and Swin Transformer~\cite{swin}.
We conduct benchmarks on ImageNet classification~\cite{imagenet}, %
COCO object detection~\cite{lin2014microsoft}, and ADE20K semantic segmentation tasks~\cite{ade20k}. %
We used min-max and LSQ~\cite{esser2019learned} quantization and reported the best performance.
We generated 10k samples with 2k optimization steps per batch with $\alpha$=1.0, and $\beta$=2.5e-5, following \citet{yin2020dreaming}.
For fine-tuning, we used $\gamma$=$\{1.0, 10.0, 100.0\}$, training for 200 epochs. 
We adapted data augmentations from SimCLR~\cite{chen2020simple} for synthetic data generation and training of \aname. 
Please refer to the supplementary material for the details.

\subsection{Comparison on Image Classification}

The experimental results are presented in \cref{tab:master}.
We first provide ``Real-Data FT'' accuracies from QAT with the original training dataset, which are considered as the empirical upper bound accuracy of DFQs. 
Then, for the DFQ designed for CNN, we utilize all components applicable to ViTs.

Overall, \aname shows significant accuracy gain in low-bit settings and various network, with a maximum gain of {51.18\%p}. %
In some cases (DeiT-S/B, Swin-B), \aname even outperforms Real-Data FT due to the proposed structural attention head distillation, which provides better guidance to follow full precision attention under a high compression rate. 
The results from the W4/A8 and W8/A8 settings show that \aname achieves similar performance compared to the Real-Data FT without access to any real samples. %

In \cref{tab:master}, the quantization accuracies of DeiT are significantly higher than those of similar-size ViTs. 
This may be due to the stronger inductive bias of DeiT compared to ViT, which enhances robustness against perturbations and preserves its capability under quantization noise.

\begin{table}[]
    \centering
   
    \resizebox{\columnwidth}{!}
    {
    \setlength{\tabcolsep}{0.8mm}
    \centering
    \begin{tabular}{clcccccccc}
    \toprule
    \multirow{3}{*}{Bits}&\multirow{3}{*}{Methods}& \multicolumn{4}{c}{COCO Dataset} & \multicolumn{4}{c}{ADE20K dataset} \\
    \cmidrule(lr){3-6} \cmidrule(lr){7-10}
     &  &   \multicolumn{2}{c}{Swin-T} & \multicolumn{2}{c}{Swin-S} & \multicolumn{4}{c}{Backbones (mIoU)} \\
    \cmidrule(lr){3-4} \cmidrule(lr){5-6} \cmidrule{7-10}
       &  & $\text{AP}_{\text{box}}$  & $\text{AP}_{\text{mask}}$ & $\text{AP}_{\text{box}}$  & $\text{AP}_{\text{mask}}$ &DeiT-S&Swin-T&Swin-S&Swin-B  \\
    \midrule
\multirow{6}{*}{\makecell{W4  \\ A4}} &	Real FT &	31.17	&	30.75	&	37.89	&	36.44	&	27.47	&	37.76	&	44.36	&	43.28	\\	\cmidrule{2-10}
&	PSAQ V1 &	\textcolor{white}00.06	&	\textcolor{white}00.06	&	\textcolor{white}00.05	&	\textcolor{white}00.06	&	\textcolor{white}00.15	&	\textcolor{white}01.65	&	\textcolor{white}03.30	&	\textcolor{white}00.89	\\	
&	PSAQ V2 &	\textcolor{white}04.52	&	\textcolor{white}05.03	&	12.12	&	12.20	&	\textcolor{white}02.60	&	\textcolor{white}03.83	&	12.13	&	\textcolor{white}06.33	\\	\cmidrule{2-10}
&	{\aname } &	\textbf{26.41}	&	\textbf{26.63}	&	\textbf{34.97}	&	\textbf{33.53}	&	\textbf{17.20}	&	\textbf{29.92}	&	\textbf{38.29}	&	\textbf{36.40}	\\	
&  Gain&		+21.89	&	+21.60	&	+22.85	&	+21.33	&	+14.60	&	+26.09	&	+26.16	&	+30.07	\\	\midrule
\multirow{6}{*}{\makecell{W5  \\ A5}} &	Real FT &	42.98	&	39.66	&	46.61	&	42.18	&	33.10	&	40.13	&	47.14	&	47.43	\\	\cmidrule{2-10}
&	PSAQ V1 &	\textcolor{white}00.41	&	\textcolor{white}00.46	&	\textcolor{white}00.64	&	\textcolor{white}00.63	&	\textcolor{white}00.80	&	20.26	&	33.10	&	39.36	\\	
&	PSAQ V2 &	32.69	&	31.21	&	45.20	&	40.99	&	\textcolor{white}05.35	&	26.35	&	37.58	&	42.01	\\	\cmidrule{2-10}
&	{\aname } &	\textbf{41.63}	&	\textbf{38.53}	&	\textbf{46.13}	&	\textbf{41.89}	&	\textbf{28.84}	&	\textbf{38.88}	&	\textbf{45.68}	&	\textbf{45.66}	\\	
& Gain &		+8.94	&	+7.32	&	+0.93	&	+0.90	&	+23.49	&	+12.53	&	+8.10	&	+3.65	\\	\midrule
\multirow{6}{*}{\makecell{W4  \\ A8}} &	Real FT &	39.55	&	38.00	&	43.34	&	41.09	&	40.96	&	42.77	&	47.56	&	47.63	\\	\cmidrule{2-10}
&	PSAQ V1 &	33.45	&	32.97	&	37.57	&	36.35	&	35.73	&	41.25	&	46.42	&	46.70	\\	
&	PSAQ V2 &	38.71	&	\textbf{37.59}	&	42.69	&	40.70	&	16.92	&	42.29	&	46.22	&	46.65	\\	\cmidrule{2-10}
&	{\aname } &	\textbf{38.77}	&	37.58	&	\textbf{42.77}	&	\textbf{40.87}	&	\textbf{41.18}	&	\textbf{43.24}	&	\textbf{46.91}	&	\textbf{47.49}	\\	
&  Gain&		+0.06	&	-0.01	&	+0.08	&	+0.17	&	+5.45	&	+0.95	&	+0.49	&	+0.79	\\	\midrule
\multirow{6}{*}{\makecell{W8  \\ A8}} &	Real FT &	46.01	&	41.63	&	48.29	&	43.13	&	41.96	&	43.62	&	46.16	&	45.39	\\	\cmidrule{2-10}
&	PSAQ V1 &	39.54	&	36.31	&	44.20	&	39.92	&	38.99	&	44.36	&	\textbf{47.68}	&	47.83	\\	
&	PSAQ V2 &	45.84	&	41.51	&	48.17	&	43.22	&	19.51	&	44.26	&	47.56	&	47.68	\\	\cmidrule{2-10}
&	{\aname } &	\textbf{46.03}	&	\textbf{41.58}	&	\textbf{48.31}	&	\textbf{43.25}	&	\textbf{41.76}	&	\textbf{44.39}	&	47.62	&	\textbf{47.87}	\\	
&  Gain&		+0.19	&	+0.07	&	+0.14	&	+0.03	&	+2.77	&	+0.03	&	-0.06	&	+0.04	\\	\bottomrule

    \end{tabular}}
        \caption{Comparison on COCO and ADE20K dataset. }

    \label{tab:master_det}
\end{table}

\subsection{Object Detection and Semantic Segmentation}

The results on the COCO object detection task in \cref{tab:master_det} show that \aname recovers from quantization error by outperforming the baseline in most settings. 
The performance gains in low-bit settings are highly noticeable, achieving up to 22.85\%p gain.
In contrast, baseline methods nearly cause the network to collapse in the W4/A4 setting. 

Results on the ADE20K semantic segmentation task also show greate improvements on low-bit settings.
The DeiT-S backbone achieves high performance gain, as it suffers from notable degradation compared to the Swin backbones.
This is because DeiTs utilize a weak inductive bias compared to Swin, which adapts architectural inductive bias. %
Therefore, DeiT backbones are vulnerable to quantization noise, as they need to preserve more information in their parameters. 

\section{Analysis}

\subsection{Sensitivity and Ablation Study }

\begin{table}[t]

     \def\arraystretch{0.7}%
    \resizebox{\columnwidth}{!}
    {
    \begin{tabular}{cccccccc}
    \toprule
       \multicolumn{3}{c}{$\loss_G$} & $\loss_T$ & \multicolumn{4}{c}{Network} \\ 
            \cmidrule(lr){1-3} \cmidrule(lr){4-4} \cmidrule(lr){5-8}
            $\mathcal{L}_{IHC}$ & $\mathcal{L}_{CL}$ & $\mathcal{L}_{TV}$ &$\mathcal{L}_{HAD}$ & ViT-T & DeiT-T & ViT-B & DeiT-B \\
    \midrule

    \xmark & \cmark & \cmark &\xmark& 13.28 & 37.70 & \textcolor{white}07.72 & 34.84 \\
    \cmark & \xmark & \cmark &\xmark&  12.31 & 16.29 & \textcolor{white}00.59 & \textcolor{white}06.78\\
    \cmark & \cmark & \xmark &\xmark&  38.38 & 50.21 & 37.80 & 56.88\\

    \midrule
    
    \cmark & \cmark & \cmark &\xmark& 39.61 & 50.16 & 39.67  & 62.11 \\
    \cmark & \cmark & \cmark &\cmark& 42.99 & 52.03 & 62.91  & 74.10 \\
    
    \bottomrule

    \end{tabular}}
        \caption{Ablation study of the loss choices.}
    \label{tab:abl}
\end{table}

\begin{table}[]
    \centering
    \def\arraystretch{0.9}%
    \resizebox{.7\columnwidth}{!}
    {
    \begin{tabular}{ccccc}
    \toprule

$\alpha$&0.01&0.1&{1}&10\\
    \midrule
ViT-T&41.41&42.54&42.67&41.04\\
ViT-B&37.09&49.87&53.96&55.32\\
DeiT-T&41.22&49.58&52.03&50.96\\
DeiT-B&54.07&60.02&63.11&63.52\\
\midrule
$\beta$&2.5E-07&2.5E-06&{2.5E-05}&2.5E-04\\
\midrule
ViT-T&41.17&41.19&42.67&41.56\\
ViT-B&48.61&48.61&53.96&51.27\\
DeiT-T&50.90&50.85&52.03&48.36\\
DeiT-B&63.77&63.13&63.11&63.13\\
\midrule
$\gamma$&0.1&{1}&{10}&{100}\\
\midrule
ViT-T&40.05&42.67&42.99&36.48\\
ViT-B&51.36&53.96&62.34&62.91\\
DeiT-T&50.65&52.03&51.01&44.34\\
DeiT-B&62.12&63.11&70.23&74.10\\
    \bottomrule
    \end{tabular}}
    \caption{Sensitivity analysis of hyperparameters. }
    \label{tab:sense}
\end{table}

We conduct an ablation study of the individual effect of loss functions, shown in \cref{tab:abl}.
Regarding synthesis loss $\loss_{G}$,
we see accuracy drops when $\loss_{CL}$ is excluded due to a lack of crucial class information from the pretrained classifier. 
As $\loss_{TV}$ only regularizes steep changes across nearby pixels, it has minor impact on quantization accuracy. 
Overall, the best results are achieved when all losses are applied in $\loss_{G}$.
Also, the proposed distillation loss $\loss_{HAD}$ boosts the quantization accuracy by up to 23.24\%p.
This method is especially effective on larger models (ViT/DeiT-B) due to their higher number of attention heads, allowing for better guidance.

\cref{tab:sense} shows the sensitivity of each hyperparameter $\alpha$, $\beta$, and $\gamma$ (\cref{eq:lg}, \cref{eq:pch2hd}), where experiment is conducted by varying each hyperparameter while others are fixed to the default value. 
The base networks perform better at higher $\alpha$, indicating that larger models can embed more class-related information in samples.
Also, models with more heads favor higher head distillation factor $\gamma$.
In contrast, $\beta$ is network-insensitive, as it only considers changes in the pixel value.

\subsection{Inter-Head Attention Similarity Metrics}
\label{sec:g_dist_acc}
\begin{table}[]
    \centering
    \setlength{\tabcolsep}{1.5mm}
    \resizebox{\columnwidth}{!}
{
    \begin{tabular}{ccccccccc}
        \toprule
 &\multicolumn{4}{c}{Generation ($g_{dist}$)}&\multicolumn{4}{c}{Distillation ($f_{dist}$)}\\
 \cmidrule(lr){2-5} \cmidrule(lr){6-9}
 \multirow{2}{*}{Network} &  \multirow{2}{*}{MSE} &  \multirow{2}{*}{\makecell{L1\\Dist.}} & \multirow{2}{*}{\makecell{KL\\Div.}} & \multirow{2}{*}{SSIM} &  \multirow{2}{*}{MSE} &  \multirow{2}{*}{\makecell{L1\\Dist.}} & \multirow{2}{*}{\makecell{KL\\Div.}} & \multirow{2}{*}{SSIM} \\ \\
        \midrule
ViT-T      & 40.91    & 40.57    & 39.63    & \textbf{42.99}   & \textcolor{white}04.71     & 37.29    & 40.38    & \textbf{42.99} \\ 
DeiT-T     & 50.88    & 50.60    & 50.33    & \textbf{52.03}  & 17.77    & 49.33    & 49.17    & \textbf{52.03}  \\ 
ViT-B      & 44.90    & 44.56    & 40.96    & \textbf{62.91}  & 23.47    & 53.25    & 59.23    & \textbf{62.91}  \\ 
DeiT-B     & {63.83}    & {63.97}    & {62.59}    & \textbf{74.10} & {68.04}    & {69.78}    & {69.59}    & \textbf{{74.10}}   \\ 
        \bottomrule
    \end{tabular}
    }
    \caption{Performance comparison on coherency metrics.}
    \label{tab:gen_dist_compare}
\end{table}

We provide a comparison of attention similarity metrics on sample synthesis and attention distillation. 
\Cref{tab:gen_dist_compare} shows that SSIM-based synthesis exhibits superior performance, while L1 distance and KL-divergence show lesser effectiveness. 
For attention-head distillation, \cref{tab:gen_dist_compare} presents similar trends that SSIM-based distillation achieves the highest accuracy, which agrees with the rank correlation coefficients in \cref{tab:att_dist_corr}.

\subsection{Computational Costs for Quantization}

\begin{figure}[]
\centering
\includegraphics[width=\columnwidth]{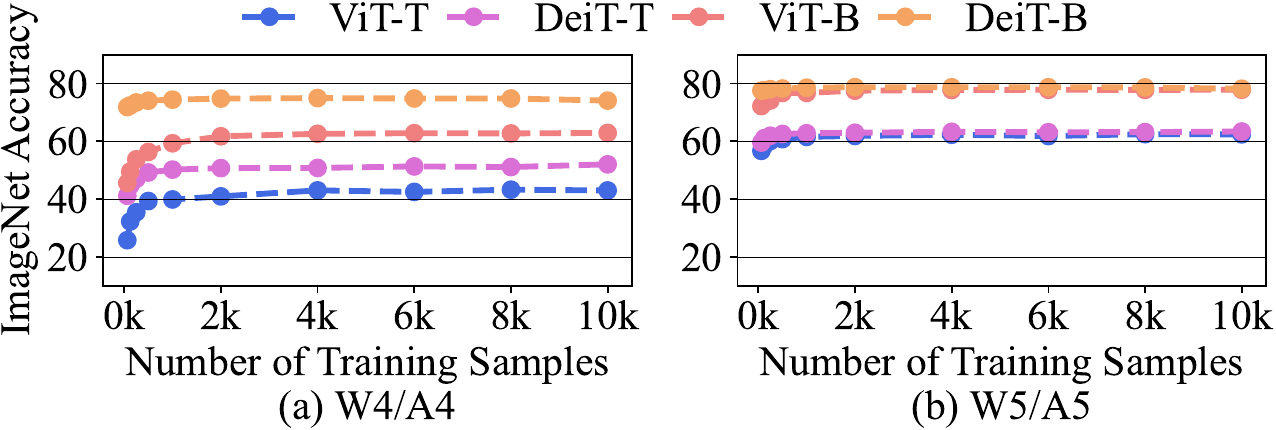}
\caption{Sensitivity analysis of synthetic dataset size.}
\label{fig:num_sample}
\end{figure}
\begin{table}[]
\centering

\resizebox{.85\columnwidth}{!}
{
\begin{tabular}{lcrrrr}
\toprule
Method & Type & Synth. & Quant. & Total & Acc. \\
\midrule
GDFQ   &QAT & -                & 10.70h  &\textcolor{red}{10.70h}                        & {11.73}        \\
AdaDFQ &QAT & -                & 8.44h& \textcolor{red}{8.44h}              & {6.21}        \\
PSAQ V1&PTQ & 0.11h             & 0.0002h&0.11h            &  0.94       \\ 
PSAQ V2&QAT & -                & 4.55h&\textcolor{blue}{4.55h}          & {2.83}      \\ \midrule
MimiQ-1k &QAT  & 1.98h             & 2.39h&\textcolor{blue}{4.37h}          & \textbf{59.32}       \\
MimiQ-4k  &QAT & 7.92h             & 2.39h&\textcolor{red}{10.31h}          & {\textbf{62.59}}        \\ 
MimiQ-10k  &QAT & 19.79h            & 2.39h&22.18h                        & \textbf{62.91}         \\
\bottomrule
\end{tabular}
}
\caption{Quantization cost comparison of ViT-B network.}
\label{tab:rebut:cost}

\end{table}

The correlation between synthetic dataset size and quantization accuracy (\cref{fig:num_sample}) reveals that \aname performs robustly with varying dataset sizes, surpassing baselines with only 64 samples in W4/A4 settings and demonstrating reasonable performance with 1k samples. 
We then compare computational costs in \cref{tab:rebut:cost}. 
In addition to the default setting of \aname using 10k samples, we also present results with similar computational costs to the baselines, where -$n$k indicates the size of the training dataset.
The comparison indicates \aname outperforms baselines with similar or fewer costs, demonstrating the cost efficiency of \aname. 

\subsection{Grad-CAM Analysis}

\begin{figure}[t]
\centering
\begin{subfigure}[t]{0.75\columnwidth}
\centering
\includegraphics[width=1.0\textwidth]{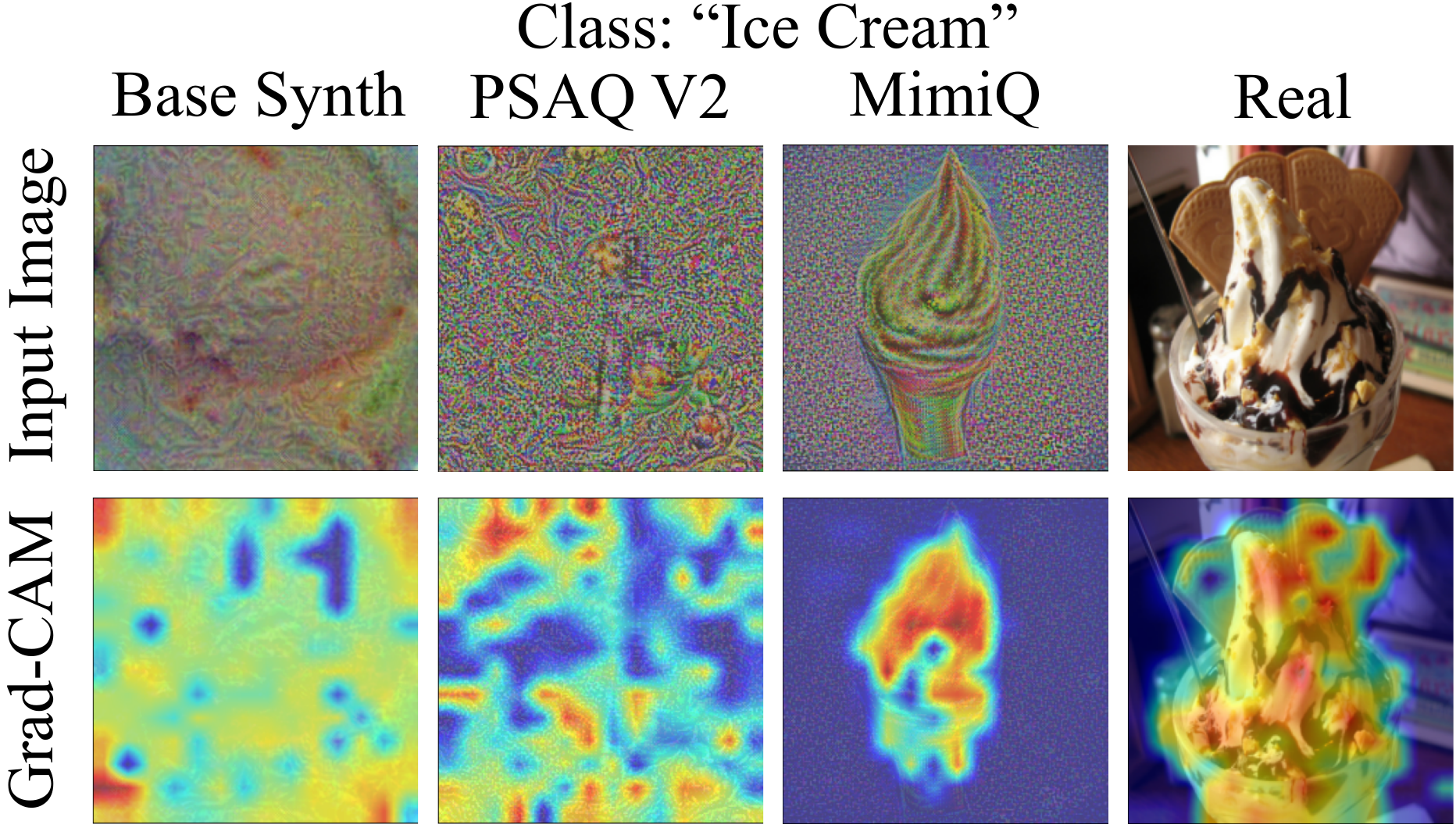}
\caption{Grad-CAM comparison of DFQ methods.}
\label{fig:grad_cam}
\end{subfigure}
\begin{subfigure}[t]{0.75\columnwidth}
\centering
\includegraphics[width=1.0\textwidth]{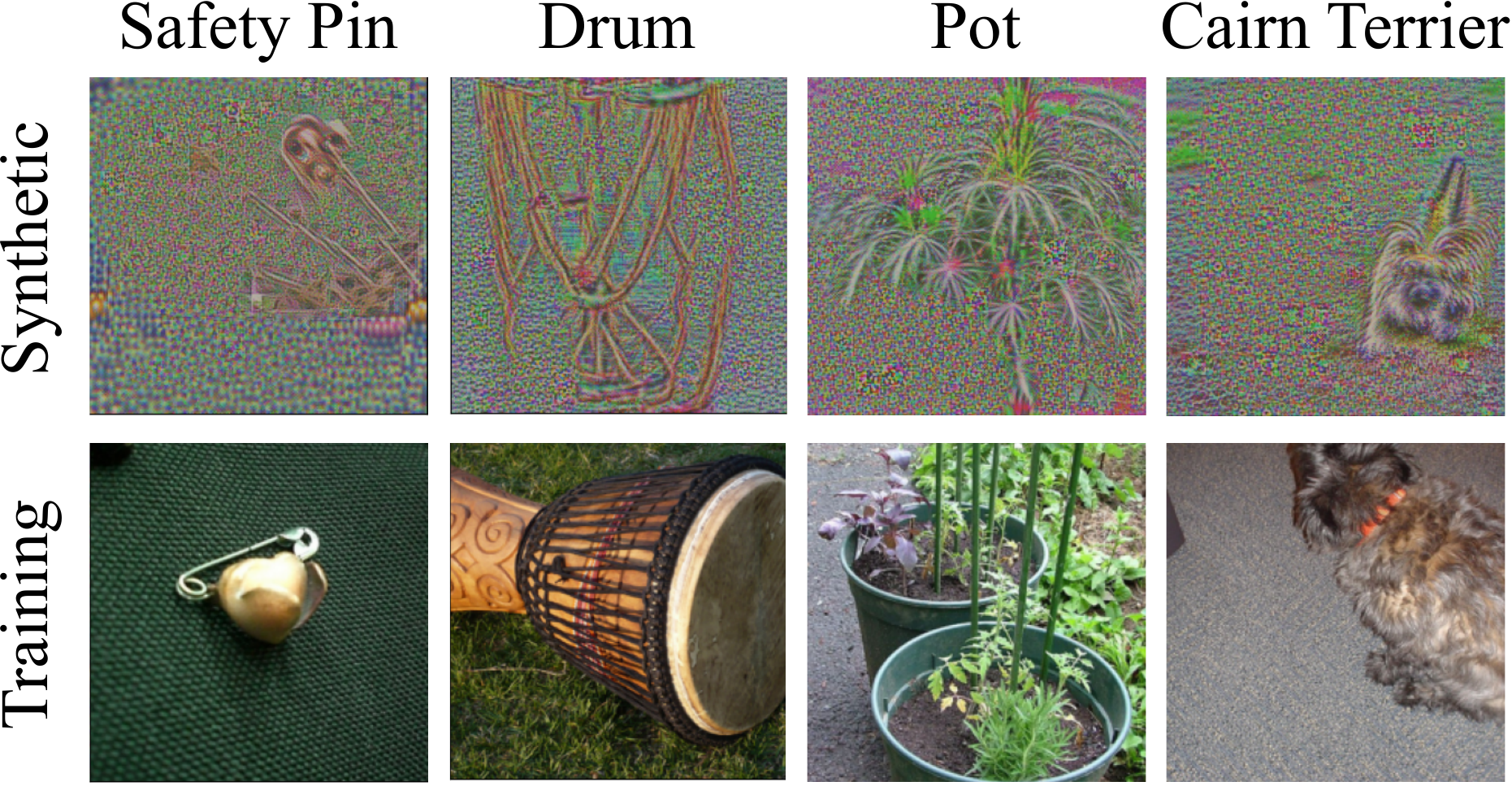}
\caption{Pairs of real/synthetic images with the lowest LPIPS. There is no indication of a privacy leak.}
\label{fig:lpips}
\end{subfigure}
\caption{
Grad-CAM and LPIPS analysis of \aname.
}
\label{fig:gradcam_lpips}
\end{figure}

As part of our analysis of how the \aname-generated images are viewed from the network's perspective, we utilize Grad-CAM~\cite{gradcam} to visualize the attention map from the last layer (\cref{fig:grad_cam}). %
It can be observed that \aname-generated images have most of their object information clear and well-clustered, similar to the real images. 
On the other hand, images from other methods have much of their object information scattered, which could harm the accuracy. %

\section{Discussion}

\begin{table}[]
    \centering

     \def\arraystretch{0.93}%
    \resizebox{\columnwidth}{!}
    {
    \centering
    \begin{tabular}{lccccc}
    \toprule
    \multirow{2}{*}{Methods} &\multirow{2}{*}{Bits} &  \multirow{2}{*}{\makecell{Calib.  data}}  &  & Models  & \\
    \cmidrule{4-6}  
       & & &  ViT-B &  DeiT-B & Swin-B \\
    \midrule
				
\multirow{4}{*}{PTQ4ViT}	&	\multirow{2}{*}{W6/A6}	&Real data &			81.65	&					80.25	&					84.01	\\										
	&		&Synth. (\aname)& 			82.40	&					80.43	&					83.96	\\	\cmidrule{2-6}									
	&	\multirow{2}{*}{W8/A8}	&Real data &			84.25	&					81.48	&					85.14	\\										
	&		&Synth. (\aname)& 			84.66	&					81.49	&					85.25	\\	\midrule									
\multirow{4}{*}{FQ-ViT}	&	\multirow{2}{*}{W8/A8/Attn4}	&Real data &			82.68	&					80.85	&					82.38	\\										
	&		&Synth. (\aname)& 			81.78	&					80.75	&					82.11	\\	\cmidrule{2-6}									
	&	\multirow{2}{*}{W8/A8/Attn8}	&Real data &			83.31	&					81.20	&					82.97	\\										
	&		&Synth. (\aname)& 			82.52	&					81.19	&					82.89	\\	\midrule									
\multirow{4}{*}{RepQ-ViT}	&	\multirow{2}{*}{W4/A4}	&Real data &			68.48	&					75.61	&					78.32	\\										
	&		&Synth. (\aname)& 			19.96	&					72.15	&					67.00	\\	\cmidrule{2-6}									
	&	\multirow{2}{*}{W6/A6}	&Real data &			83.62	&					81.27	&					84.57	\\										
	&		&Synth. (\aname)& 			82.35	&					80.84	&					82.36	\\											
    \bottomrule
    \end{tabular}}
        \caption{Comparison with real-data ViT quantizations.}
    \label{tab:advquant}
        \vspace{-1mm}
\end{table}

\textbf{Adaptation to Real-Data ViT Quantization.} 
To further explore the efficacy of \aname, we benchmark adaptation of \aname samples to real-data ViT quantization methods (PTQ4ViT, FQ-ViT, RepQ-ViT) following the original settings of each paper.
The results are shown in \cref{tab:advquant}, where %
``Attn$k$'' refers to $k$-bit attention quantization. %
Please note that these experiments utilize only synthetic samples from the \aname framework, replacing the proposed attention distillation phase with existing quantization methods. 
In most cases, calibration with \aname samples achieves an accuracy similar to that of real-data calibration. 
Notably, our results on PTQ4ViT show that \aname often outperforms the original results, suggesting \aname's potential to enhance real-data quantization. 
However, we observe accuracy drops in RepQ-ViT in the W4/A4 setting. %
This may be due to the overfitting to synthetic samples leading to distribution shifts, where \aname framework counters this by minimizing head-wise attention discrepancy.
Overall, the results demonstrate \aname's versatility and its potential for broader application.

\begin{table}
\centering
    \resizebox{0.75\columnwidth}{!}
    {
    \begin{tabular}{lcc}
        \toprule
        \multirow{2}{*}{\makecell{ImageNet\\Acc. (\%)}} & Synthetic/Real & Synthetic$\rightarrow$Real \\
        &  Distinguishability & Transferability \\
        \midrule
        Train & 99.97 & 49.69 \\
        Test & 99.99 & \textcolor{white}00.16 \\
        \bottomrule
    \end{tabular}
    }
    \caption{Experiments on model inversion and identity attacks using ResNet-18. Results indicate the attacks fail.}
    \label{tab:attack}
\end{table}

\subsubsection{Does \aname Threaten Privacy?}
\aname may be linked to input reconstruction attacks~\cite{reconst1, reconst2} that resemble specific images of original data.
Comparison using LPIPS against the original training dataset reveals that only general features are captured without replicating specific images (\cref{fig:lpips}). 
Further tests following \citeauthor{prakash2020privacy} (\cref{tab:attack}) shows that \aname samples are distinguishable from real ones with near-perfect train and test accuracies of 99.97\% and 99.99\%, mitigating identity attack concerns. 
Finally, a synthetic-trained network from scratch underperformed dramatically (0.16\%), suggesting a low risk of model inversion attacks.

\section{Conclusion}
In this paper, we propose \aname, a DFQ for ViTs inspired by attention similarity.
We observe head-wise attention maps of synthetic samples are not aligned and aligning them contributes to the quantization accuracy.
From the findings, \aname utilizes inter-head attention similarity to better leverage the knowledge instilled in the attention architecture, synthesizing training data that better aligns inter-head attention. 
In addition, \aname utilizes fine-grained head-wise attention map distillation. %
As a result, \aname brings significant performance gain, setting a new state-of-the-art results.

\section{Acknowledgments}

This work was partially supported by 
National Research Foundation of Korea (NRF) grant funded by the Korea government (MSIT) (No.2022R1C1C1011307), %
Institute of Information \& communications Technology Planning \& Evaluation (IITP) (RS-2021II211343 (SNU AI), RS-2022-II220871, RS-2021-II212068 (AI Innov. Hub)), 
and an unrestricted gift from Google.
Disclaimer:  Any opinions, findings, and conclusions or recommendations expressed in this material are those of the author(s) and do not necessarily reflect the views of Google.

\bibliography{aaai25}

\clearpage

\appendix
\begin{strip}
\centering
{\LARGE\bfseries Supplementary Materials}
\end{strip}
\twocolumn

\section{Detailed Experimental Settings}
\label{sec:supp:expr}

We used the Timm library~\cite{timm} for network implementation and pretrained weights to benchmark on ImageNet~\cite{imagenet} classification tasks.
In addition, the implementations and pretrained weights for object detection and semantic segmentation are from the mmdetection~\cite{mmdetection} and mmsegmentation~\cite{mmseg2020}, respectively. %
We used the Adam optimizer for the synthetic image generation with lr=0.1 and $b_1$=0.9, $b_2$=0.999, a batch size of 32.
For fine-tuning quantized networks, we used the SGD optimizer with Nesterov momentum of 0.9 with lr=1e-3, batch size of 16, and $\gamma$=$\{1.0, 10.0, 100.0\}$, training for 200 epochs.

\section{Details of Quantization}
\label{sec:supp:quant}

We use a uniform quantization method shown in \cref{eq:q_basic} for all experimental settings. 
We use per-channel symmetric quantization for weight parameters where $z$ is $\mathbf{0}$, and each output channel uses different $s$. 
For activations, we use per-tensor asymmetric quantization which uses a single ($s$, $z$) pair for the output tensor. 

To compute quantization parameters $s$ and $z$, we use the min-max quantization and LSQ~\cite{esser2019learned}. 
The min-max quantization determines quantization parameters from minimum and maximum values from the target tensor as follows:
\begin{align}
    s &= \nicefrac{(2^k-1)}{(max(\theta)-min(\theta))}, \label{eq:asymq1}\\
    z &= s \times min(\theta) + 2^{k-1}. \label{eq:asymq2} 
\end{align}
For activation quantization, $max(\theta)$ and $min(\theta)$ are determined by the exponential moving average of activation from each iteration. 
LSQ initializes ($s$, $z$) with ($s$,$z$) from \cref{eq:asymq1} and \cref{eq:asymq2}, then $s$ and $z$ are updated via gradient descent with the loss function $\mathcal{L_{T}}$, which is used to train a quantized network. 
As the gradients of ($s$, $z$) are accumulated from each parameters of $\theta$, LSQ scales gradients $\nabla_{s} \mathcal{L}$ by $\nicefrac{1}{\sqrt{numel(\theta) \cdot q_{max}}}$, where $numel(\theta)$ is the number of elements in $\theta$.
Note that this difference between the min-max and LSQ quantization is only in the training phase; they have identical computations in the quantized inference.

\begin{figure}[t]
\centering
\begin{subfigure}[t]{0.48\columnwidth}
\centering
\includegraphics[width=1.0\textwidth]{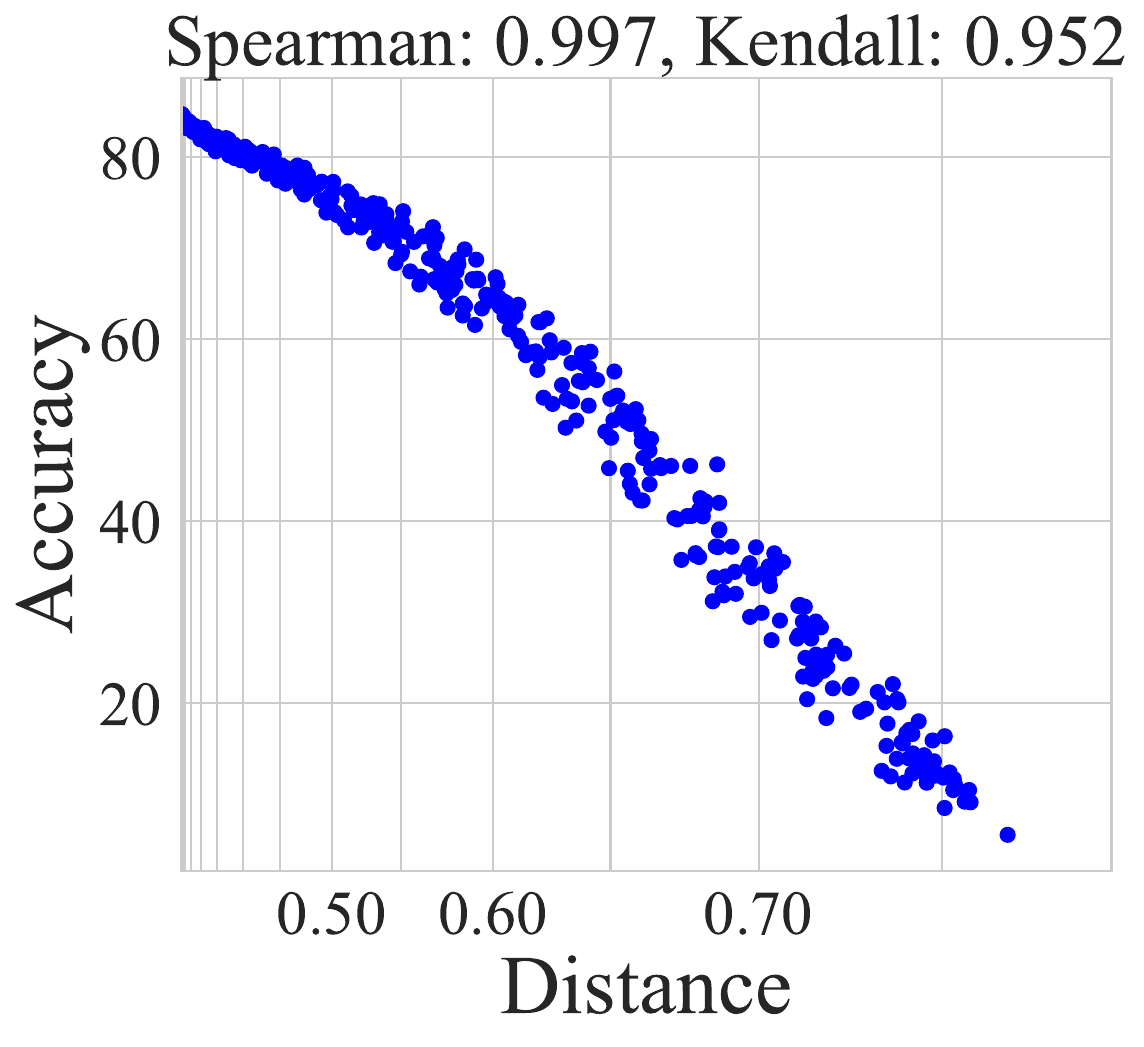}
\caption{DSSIM}
\label{fig:corr:dssim}
\end{subfigure}
\begin{subfigure}[t]{0.48\columnwidth}
\centering
\includegraphics[width=1.0\textwidth]{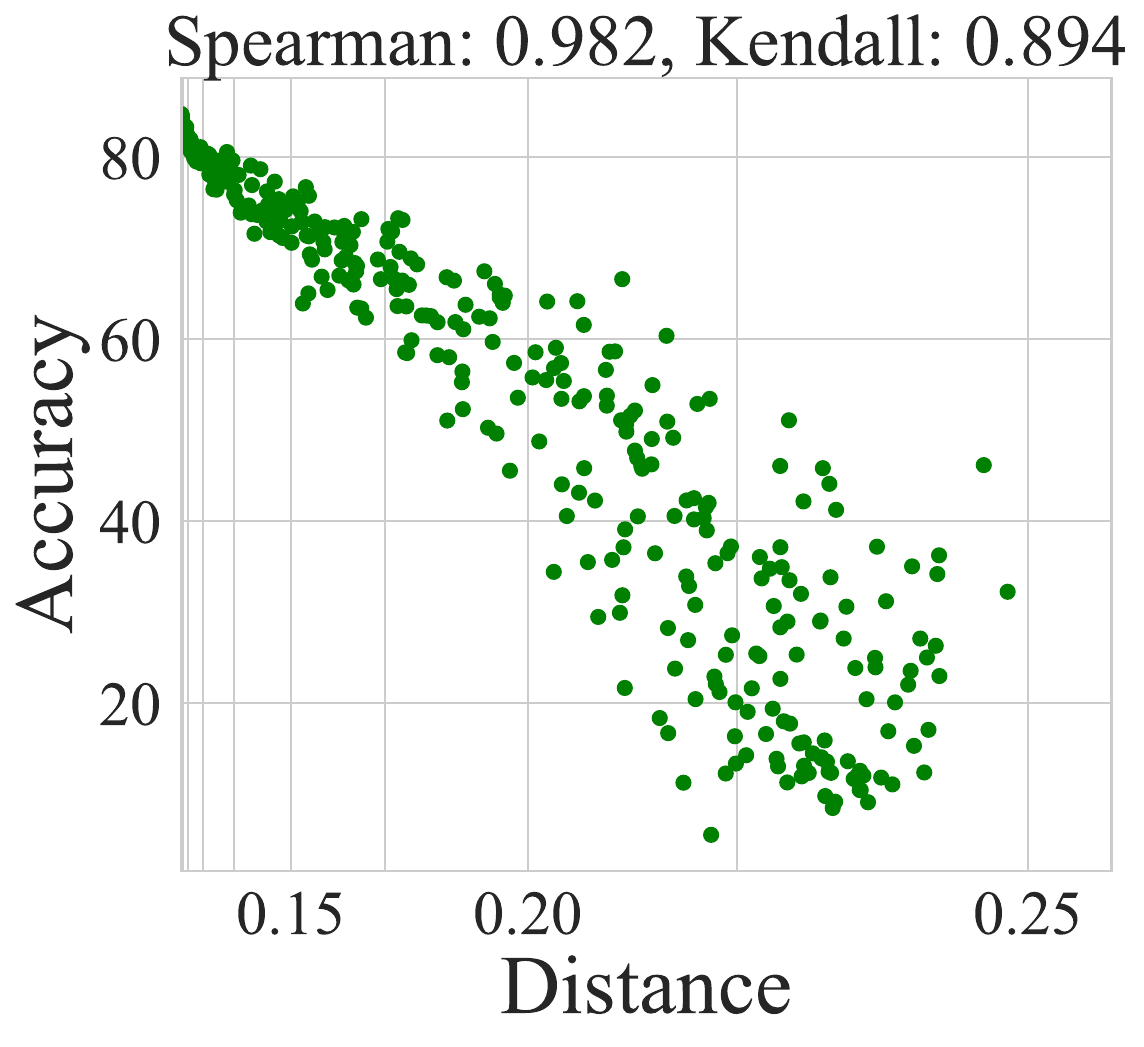}
\caption{MSE}
\label{fig::corr:mse}
\end{subfigure}
\begin{subfigure}[t]{0.48\columnwidth}
\centering
\includegraphics[width=1.0\textwidth]{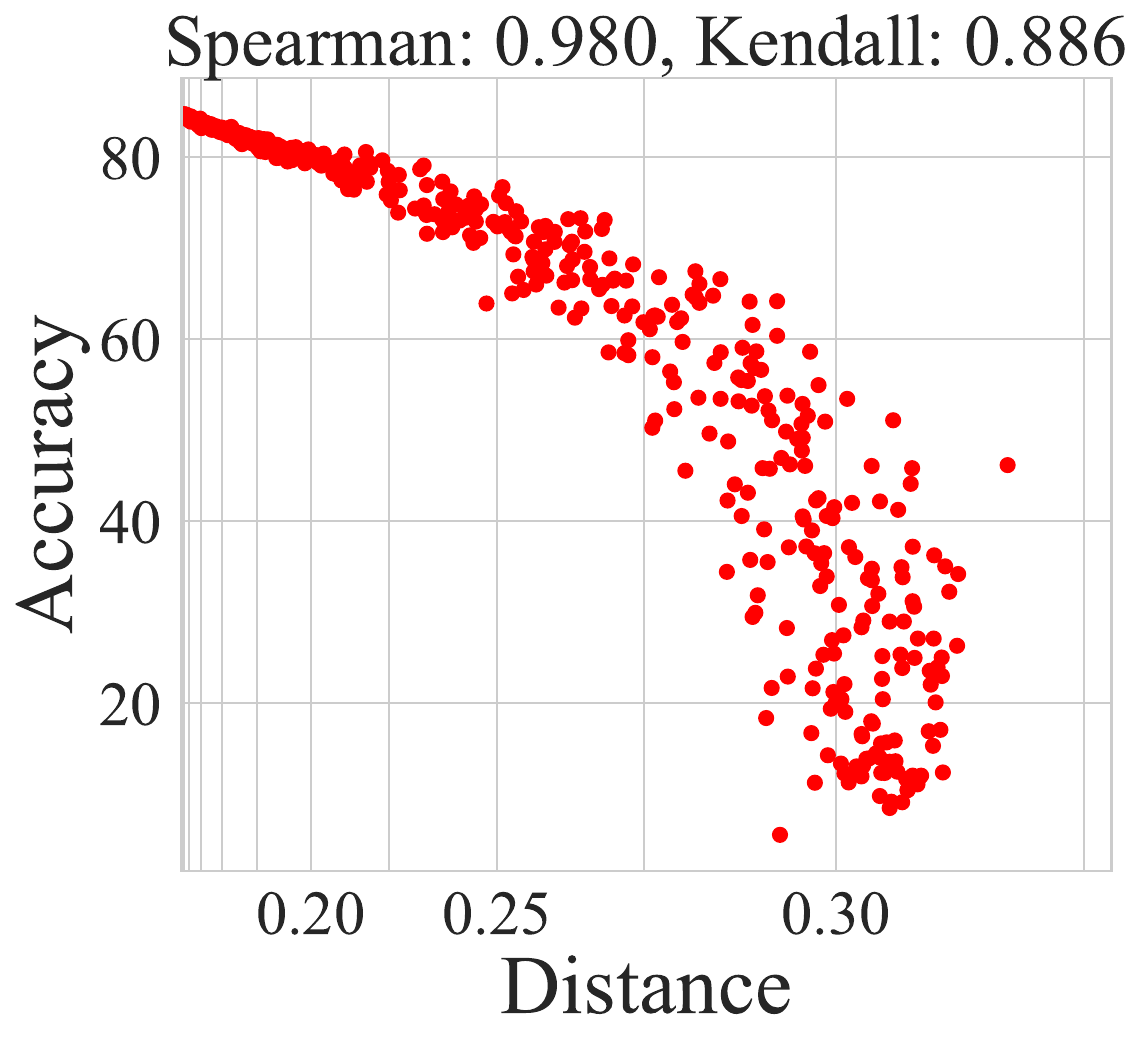}
\caption{L1-Distance}
\label{fig:corr:l1}
\end{subfigure}
\begin{subfigure}[t]{0.48\columnwidth}
\centering
\includegraphics[width=1.0\textwidth]{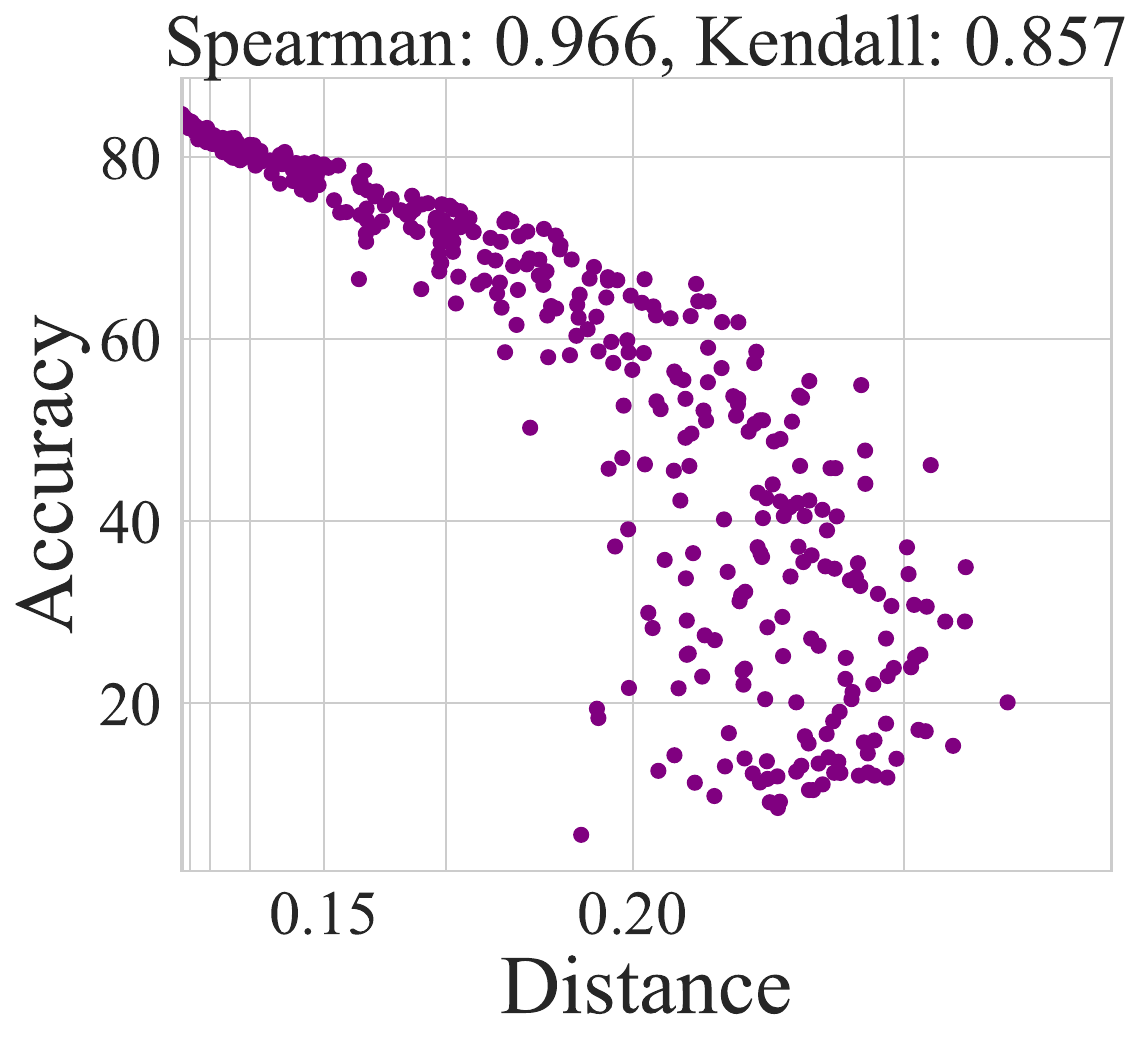}
\caption{KL-Divergence}
\label{fig:corr:kl}
\end{subfigure}
\caption{This figure illustrates the comparison of four different attention coherency metrics: (a) DSSIM, (b) MSE, (c) L1 Distance, and (d) KL-Divergence to the network accuracy. DSSIM exhibits consistent trends with accuracy, indicating a stronger rank correlation compared to the other metrics.  The results highlight the significance of choosing appropriate attention similarity metrics.}
\label{fig:corr_vis}
\end{figure}

\section{Visualization of Coherency-Accuracy Correlation}
\label{sec:supp:corr}

We visualize the correlation between mean attention map distance with respect to the full-precision model, using each distance metric.
The visualization results are shown in \cref{fig:corr_vis}.
Each plot visualizes 500 settings sampled from the configuration space, where attention heads are randomly quantized layer-wise. 
Among various distance metrics, \cref{fig:corr:dssim} exhibits a tightly packed cluster that indicates its strong correlation for accuracy.
While the other metrics also reveal some correlation between accuracy, they show a broader dispersion of data points reflecting a weaker correlation. 
The visualization results further support our claim that DSSIM can be a robust proxy of quantized accuracy.

\begin{figure}
\centering
\includegraphics[width=\columnwidth]{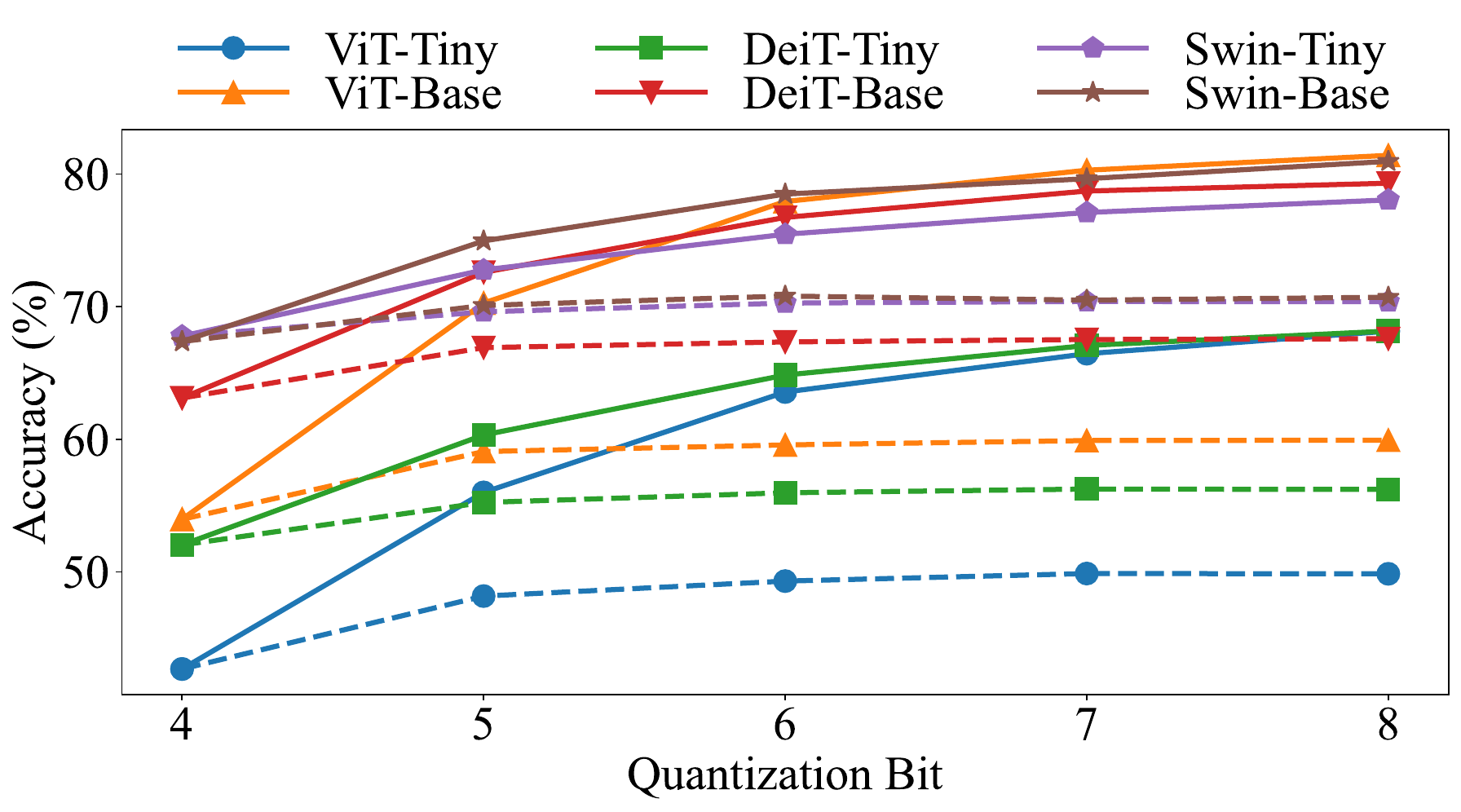}
\hspace{0.2cm}
\caption{Sensitivity analysis on quantization bitwidth of weight and activation. The solid line depicts the model accuracy varying activation bitwidth, while the dotted line indicates the model accuracy varying weight parameter bitwidth. Different colors represent distinct ViT architectures. Best viewed in color. }
\label{fig:supp:quant_sense}
\end{figure}

\section{Sensitivity Analysis on Weight and Activation Quantization Bitwidths}

In this section, we provide a sensitivity analysis of ViT architectures focusing on the weight and activation bitwidths of the quantized ViT model.
Specifically, we set the W4/A4 quantization setting as a starting point and investigate the impact of incrementally varying each quantization bitwidth of weight and activation. 

The results of this sensitivity analysis are depicted in \cref{fig:supp:quant_sense}.
The solid lines represent the accuracy sensitivity of activation quantization, while dotted lines correspond to the weight quantization sensitivity. 
The same line color indicates the same ViT architecture. 

A key insight from this analysis is that the quantized ViT architecture is more sensitive to the activation bitwidth, while increasing weight parameter bitwidth gives relatively small effects on the quantization accuracy. 
The results show the relative importance of activation quantization in ViT model compression, indicating the need for proper activation quantization for better quantization accuracy.

\section{More Visualization Result of Head-wise Attention Map}
\label{sec:supp:more_vis}

\begin{figure*}[t]
\centering
\begin{subfigure}[]{0.99\textwidth}
\centering
\includegraphics[width=0.93\textwidth]{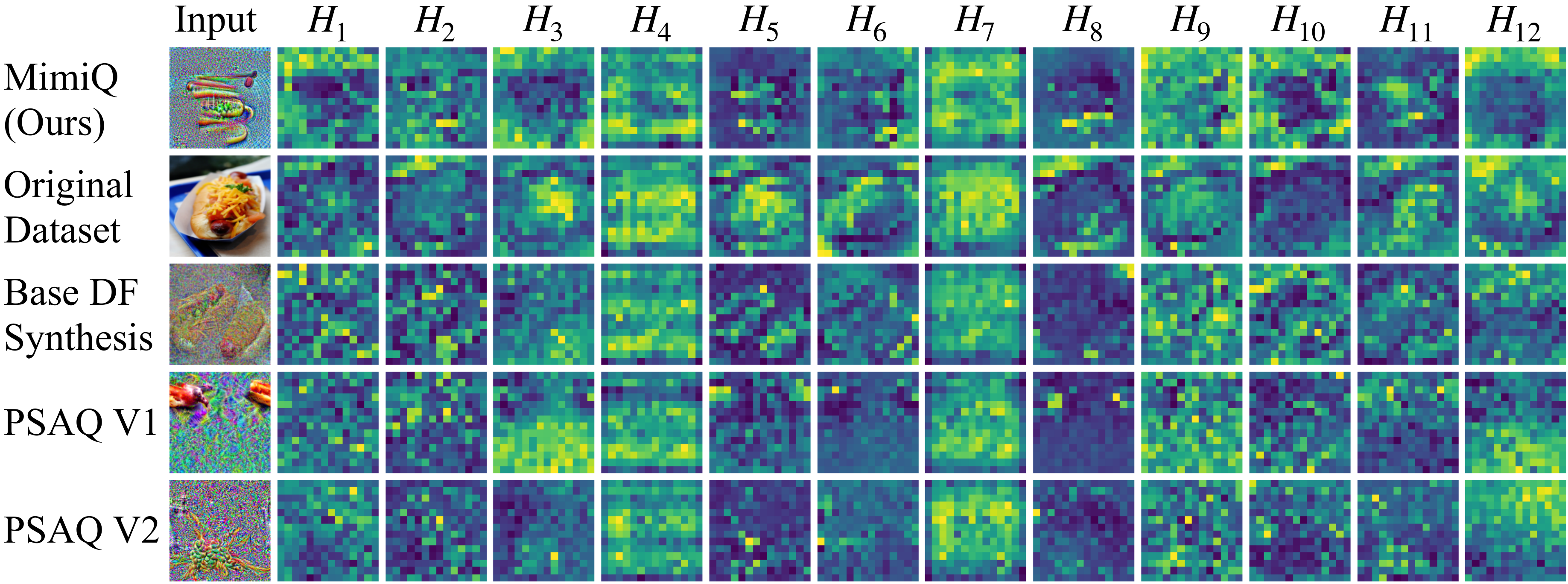}
\vspace{3mm}
\end{subfigure}
\begin{subfigure}[]{0.99\textwidth}
\centering
\includegraphics[width=0.93\textwidth]{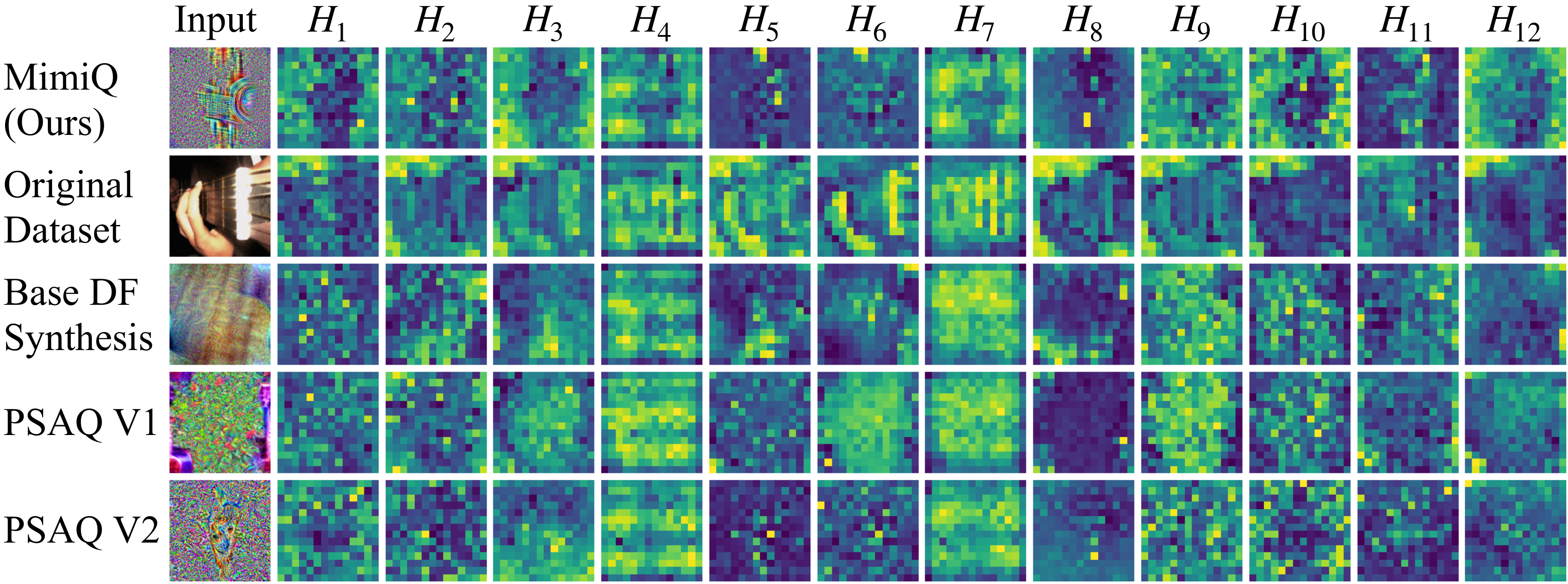}
\vspace{3mm}
\end{subfigure}
\begin{subfigure}[]{0.99\textwidth}
\centering
\includegraphics[width=0.93\textwidth]{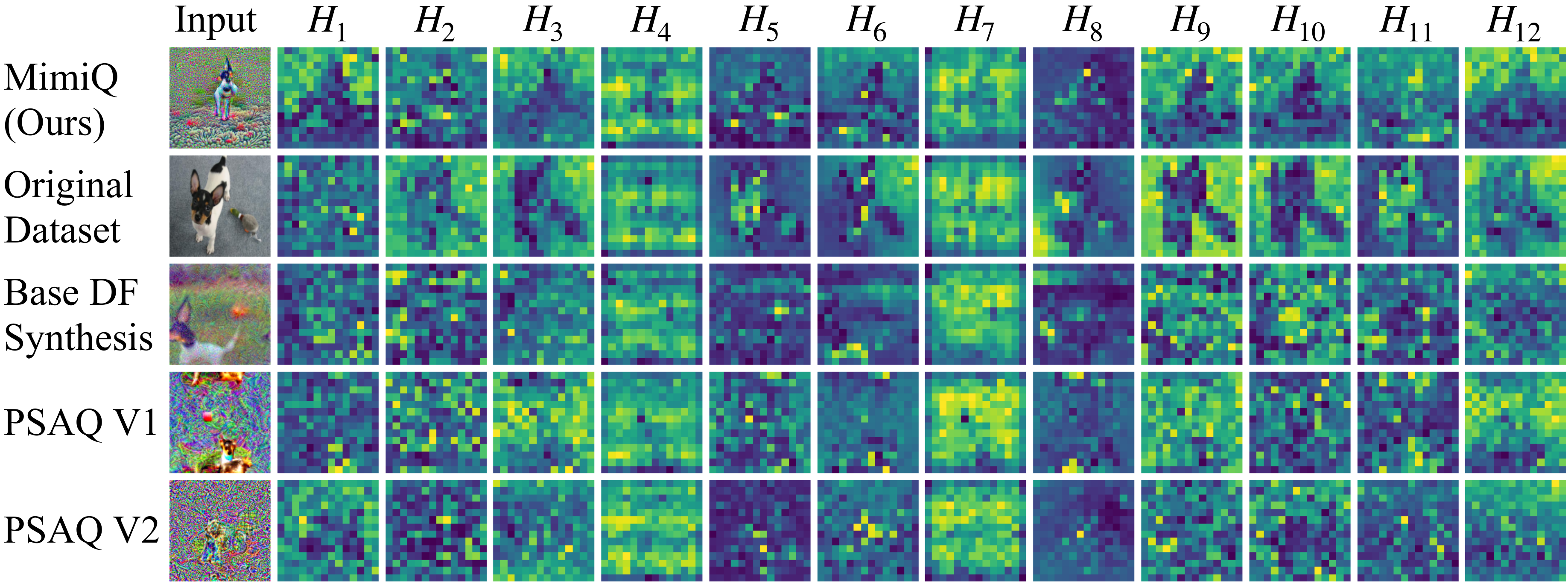}
\end{subfigure}

\caption{Further attention map visualization of the input image from \aname, Real training samples, and DFQ baselines. The attention map is generated from \textbf{ViT-Base} architecture. Each row represents different input images and each column represents one of the attention heads. }
\label{fig:synth_compare_vb}
\end{figure*}

\begin{figure*}[t]
\centering
\begin{subfigure}[t]{0.99\textwidth}
\centering
\includegraphics[width=0.93\textwidth]{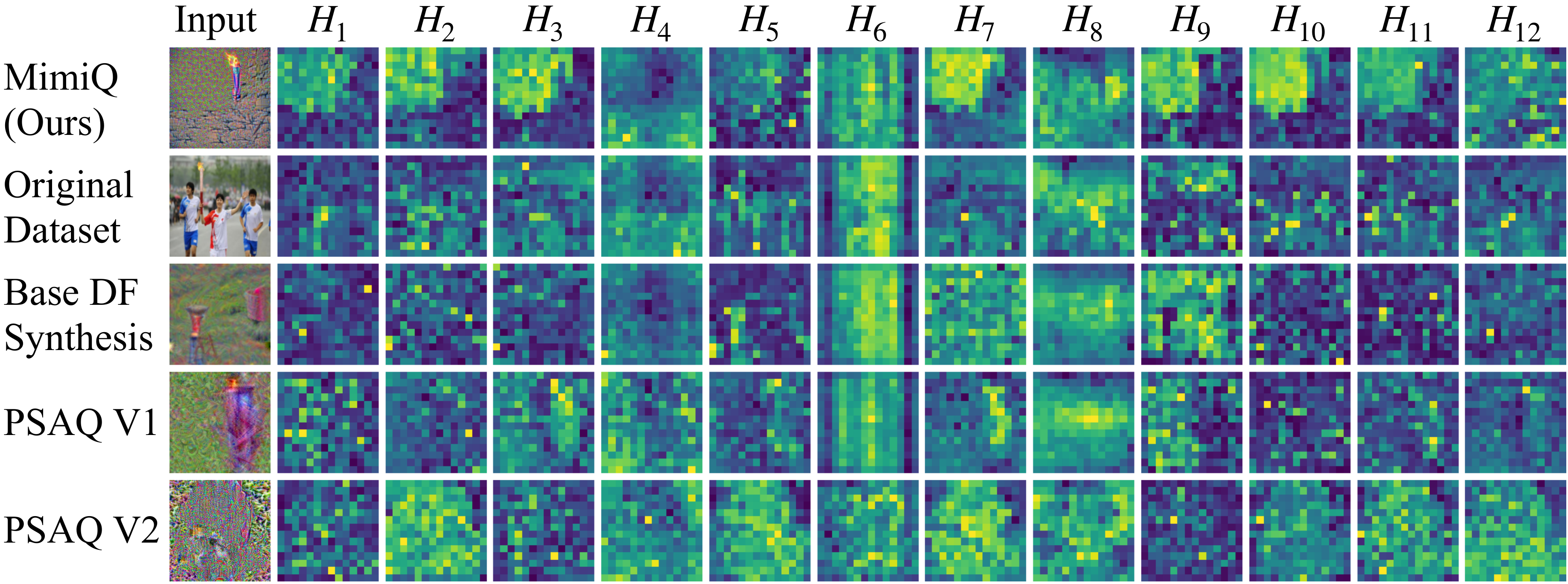}
\vspace{3mm}
\end{subfigure}
\begin{subfigure}[t]{0.99\textwidth}
\centering
\includegraphics[width=0.93\textwidth]{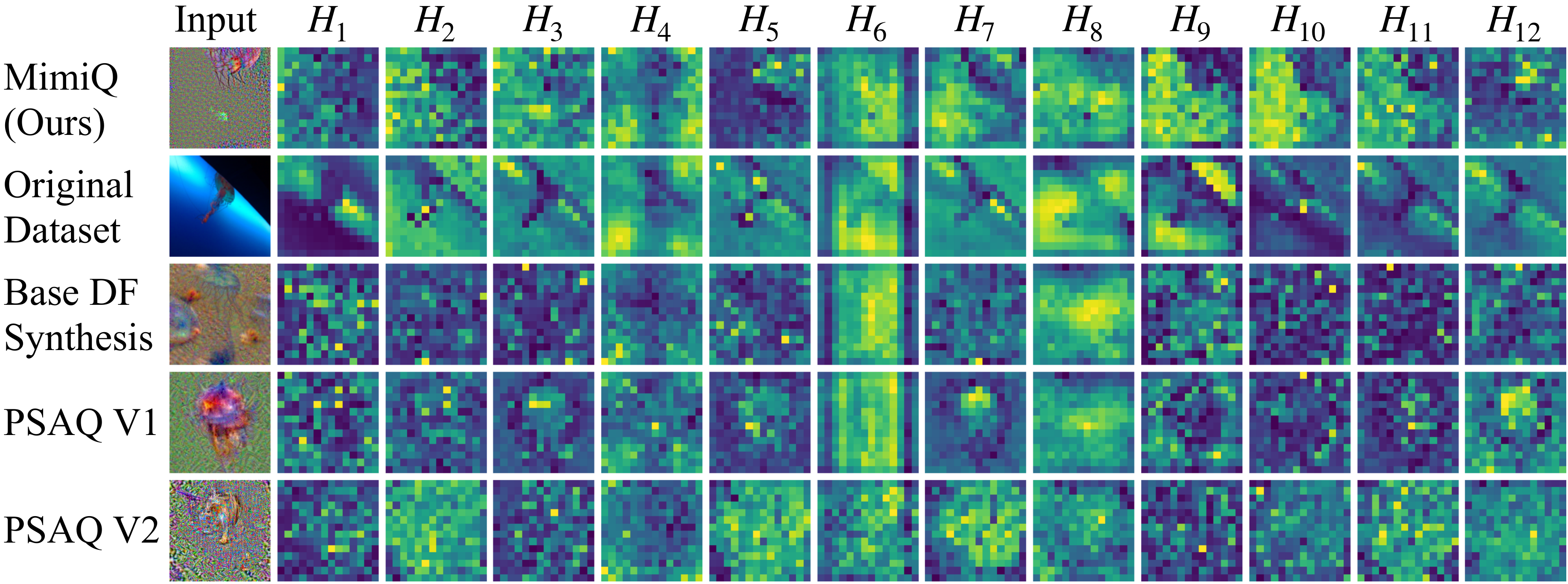}
\vspace{3mm}
\end{subfigure}
\begin{subfigure}[t]{0.99\textwidth}
\centering
\includegraphics[width=0.93\textwidth]{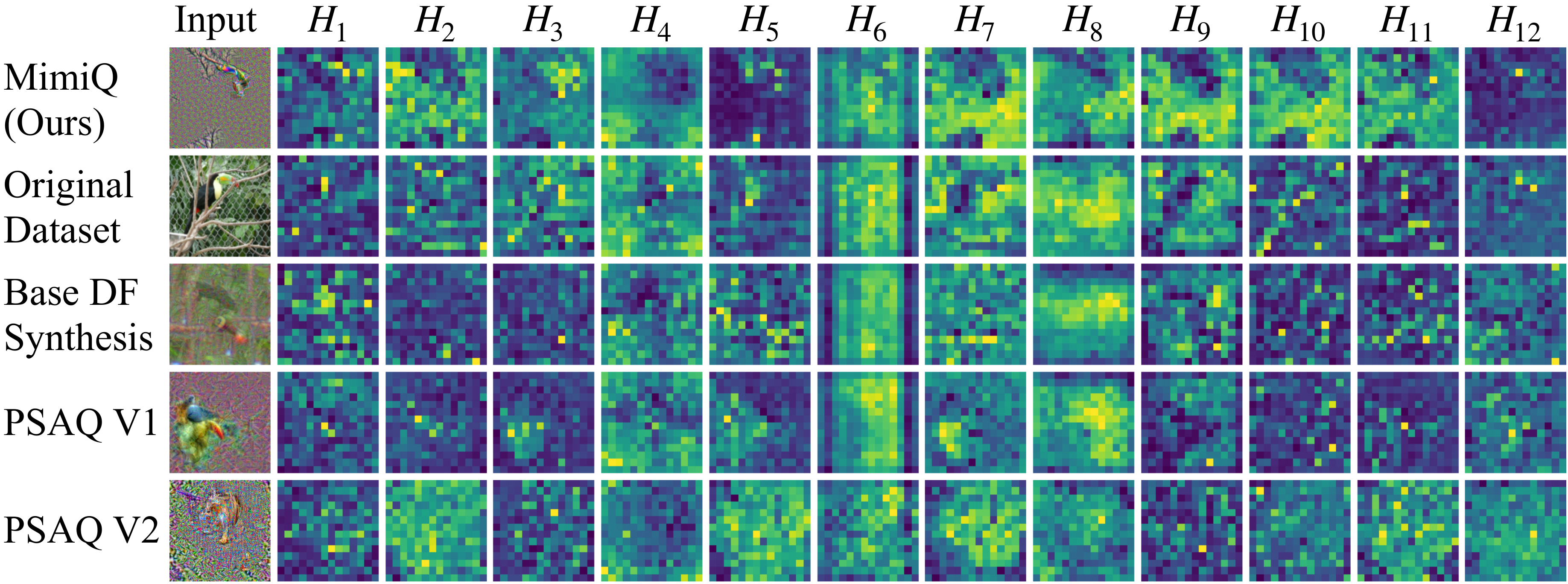}
\end{subfigure}

\caption{Further attention map visualization of the input image from \aname, Real training samples, and DFQ baselines. The attention map is generated from \textbf{DeiT-Base} architecture. Each row represents different input images and each column represents one of the attention heads. }
\label{fig:synth_compare_db}
\end{figure*}

\begin{figure*}[t]
\centering
\begin{subfigure}[t]{0.6\textwidth}
\centering
\includegraphics[width=0.93\textwidth]{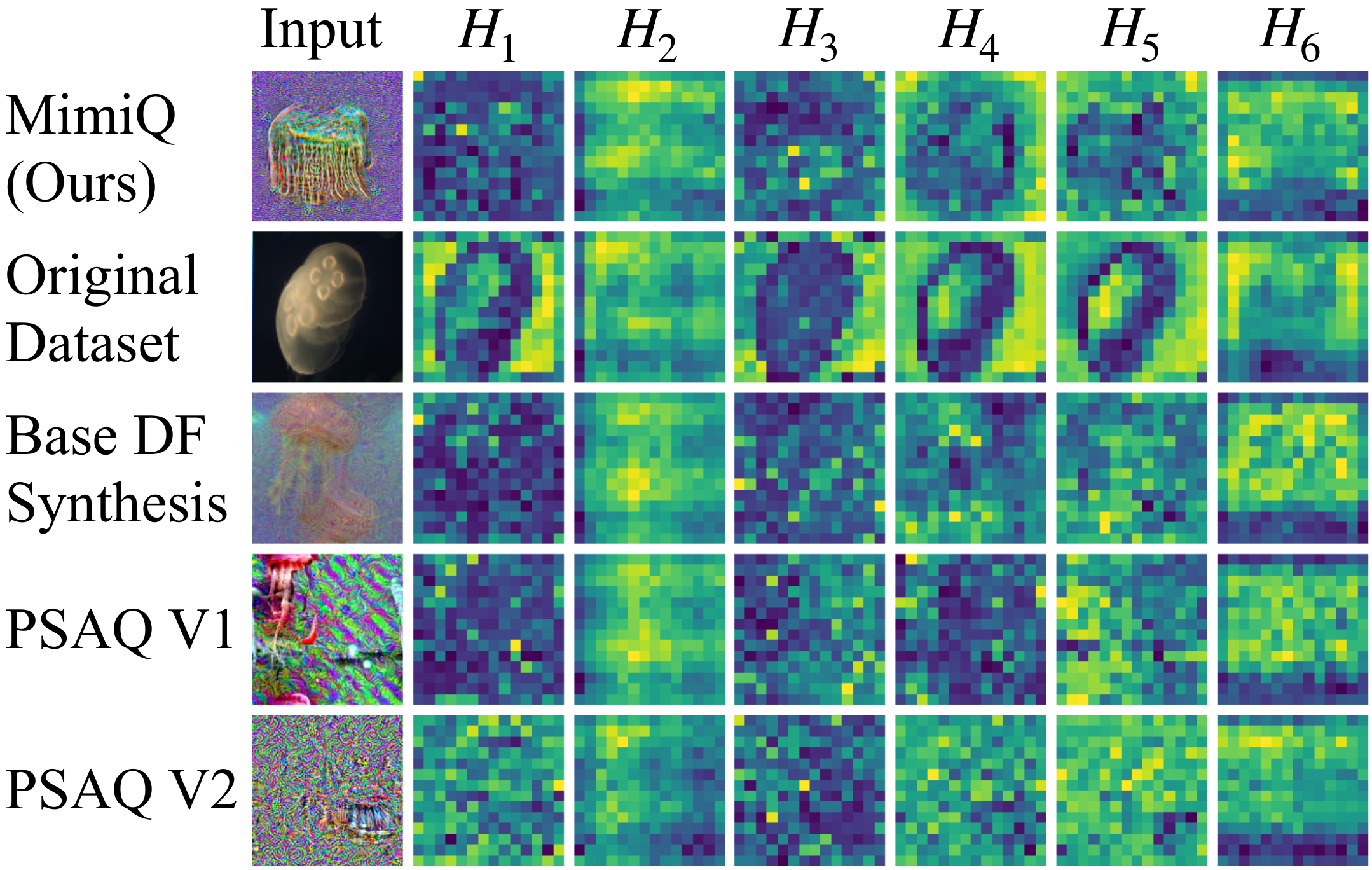}
\vspace{3mm}
\end{subfigure}
\begin{subfigure}[t]{0.6\textwidth}
\centering
\includegraphics[width=0.93\textwidth]{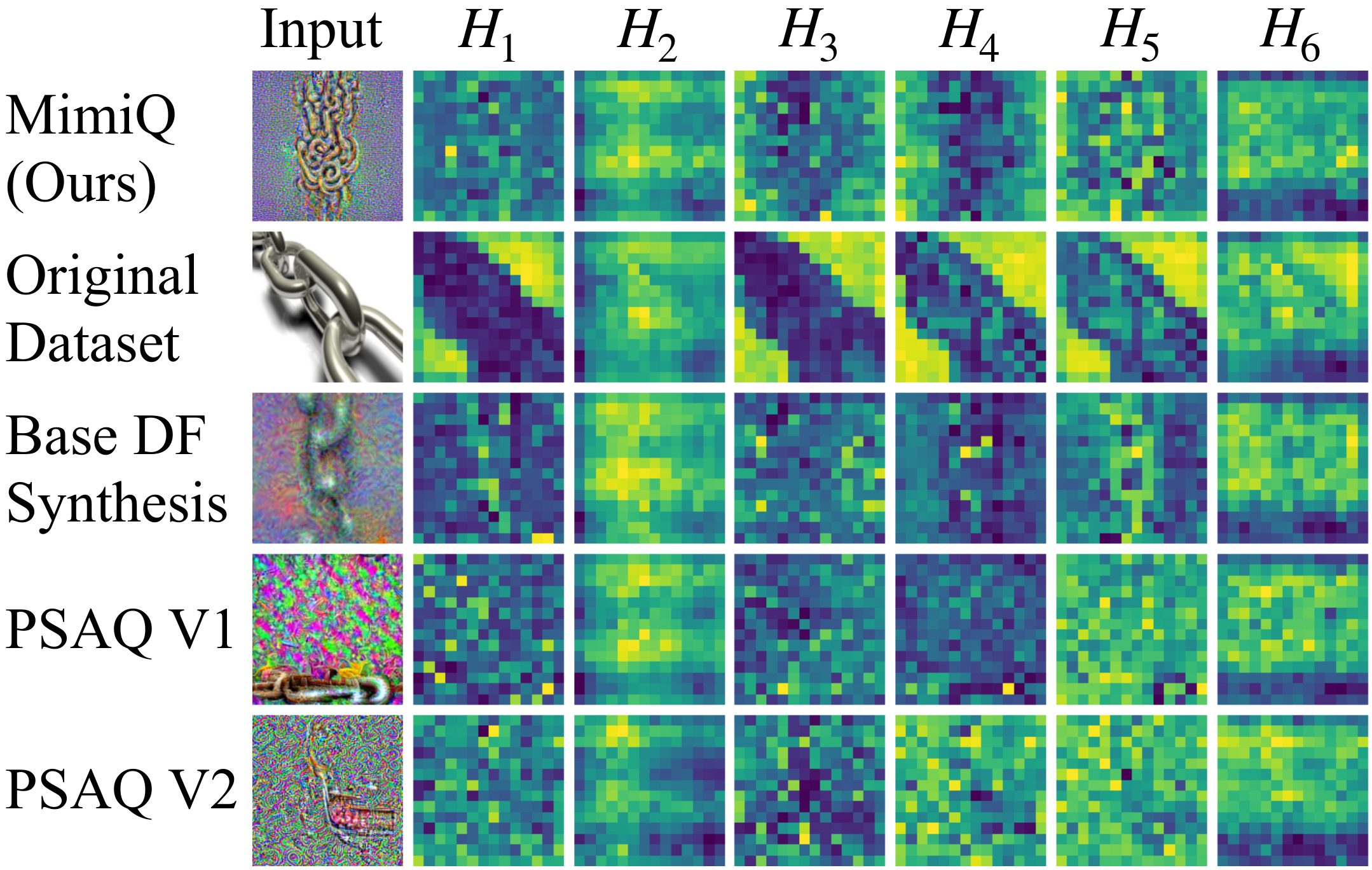}
\vspace{3mm}
\end{subfigure}
\begin{subfigure}[t]{0.6\textwidth}
\centering
\includegraphics[width=0.93\textwidth]{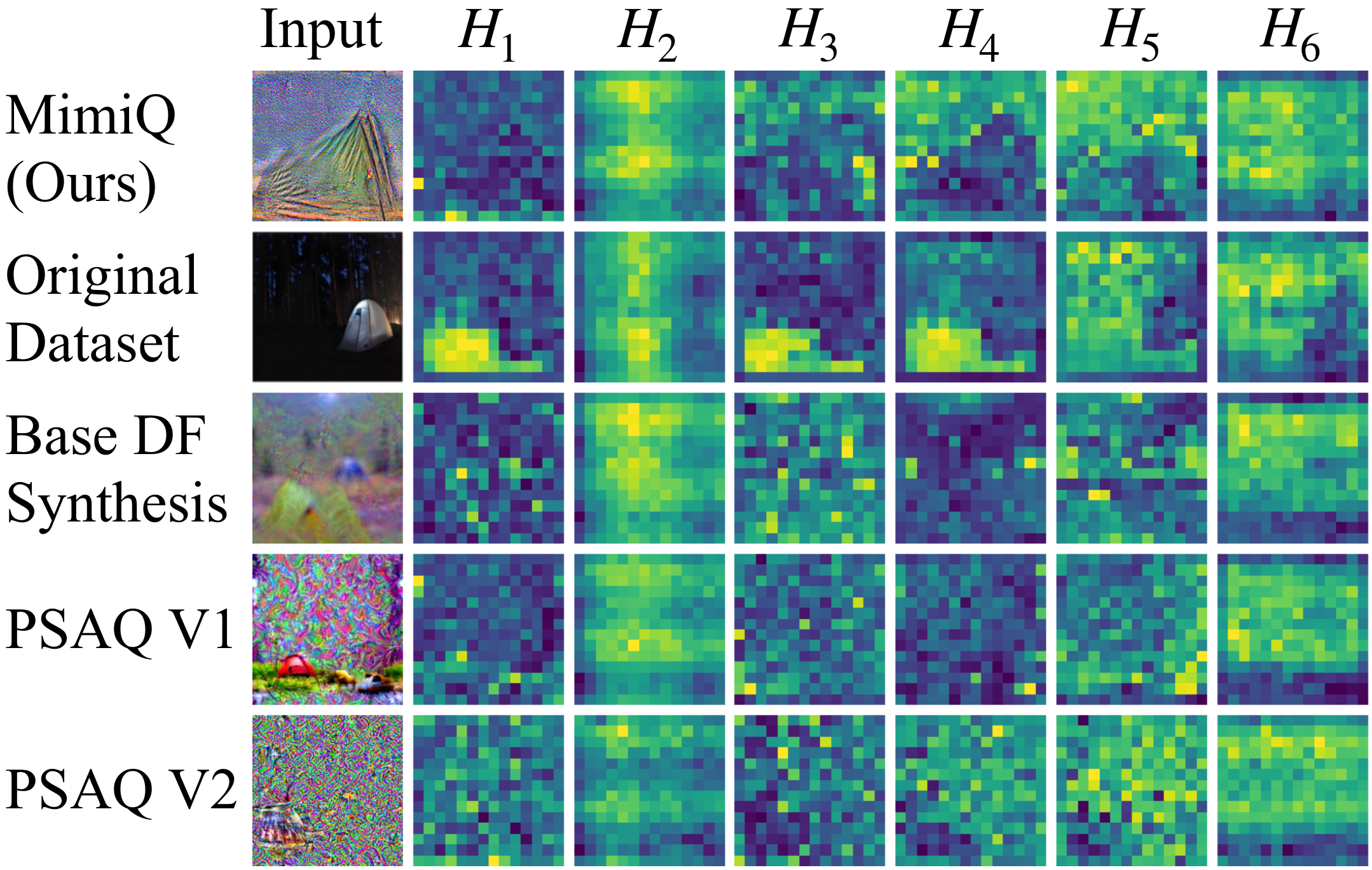}
\end{subfigure}

\caption{Further attention map visualization of the input image from \aname, Real training samples, and DFQ baselines. The attention map is generated from \textbf{ViT-Small} architecture. Each row represents different input images and each column represents one of the attention heads. }
\label{fig:synth_compare_vs}
\end{figure*}

\begin{figure*}[t]
\centering
\begin{subfigure}[t]{0.6\textwidth}
\centering
\includegraphics[width=0.93\textwidth]{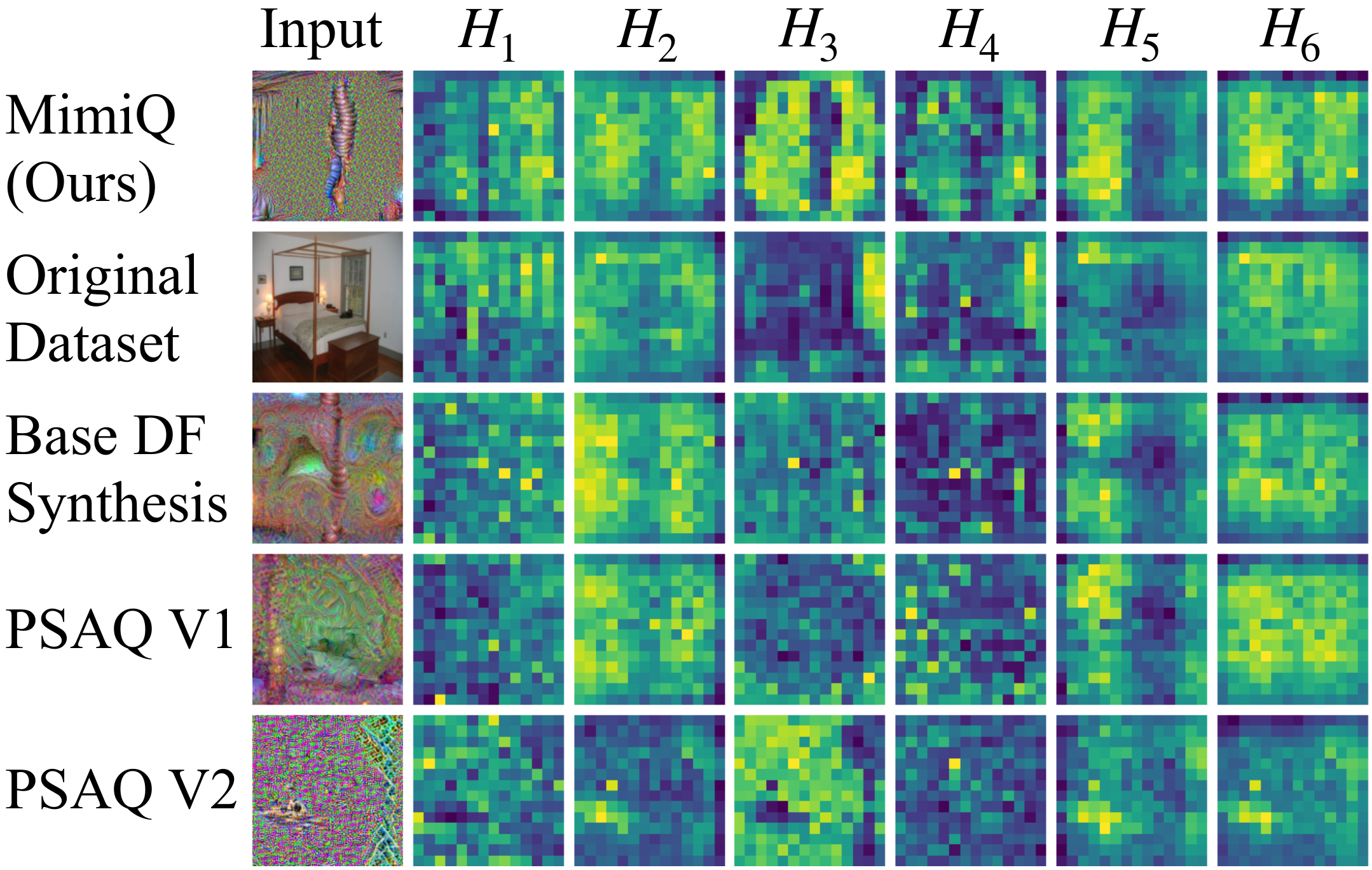}
\vspace{3mm}
\end{subfigure}
\begin{subfigure}[t]{0.6\textwidth}
\centering
\includegraphics[width=0.93\textwidth]{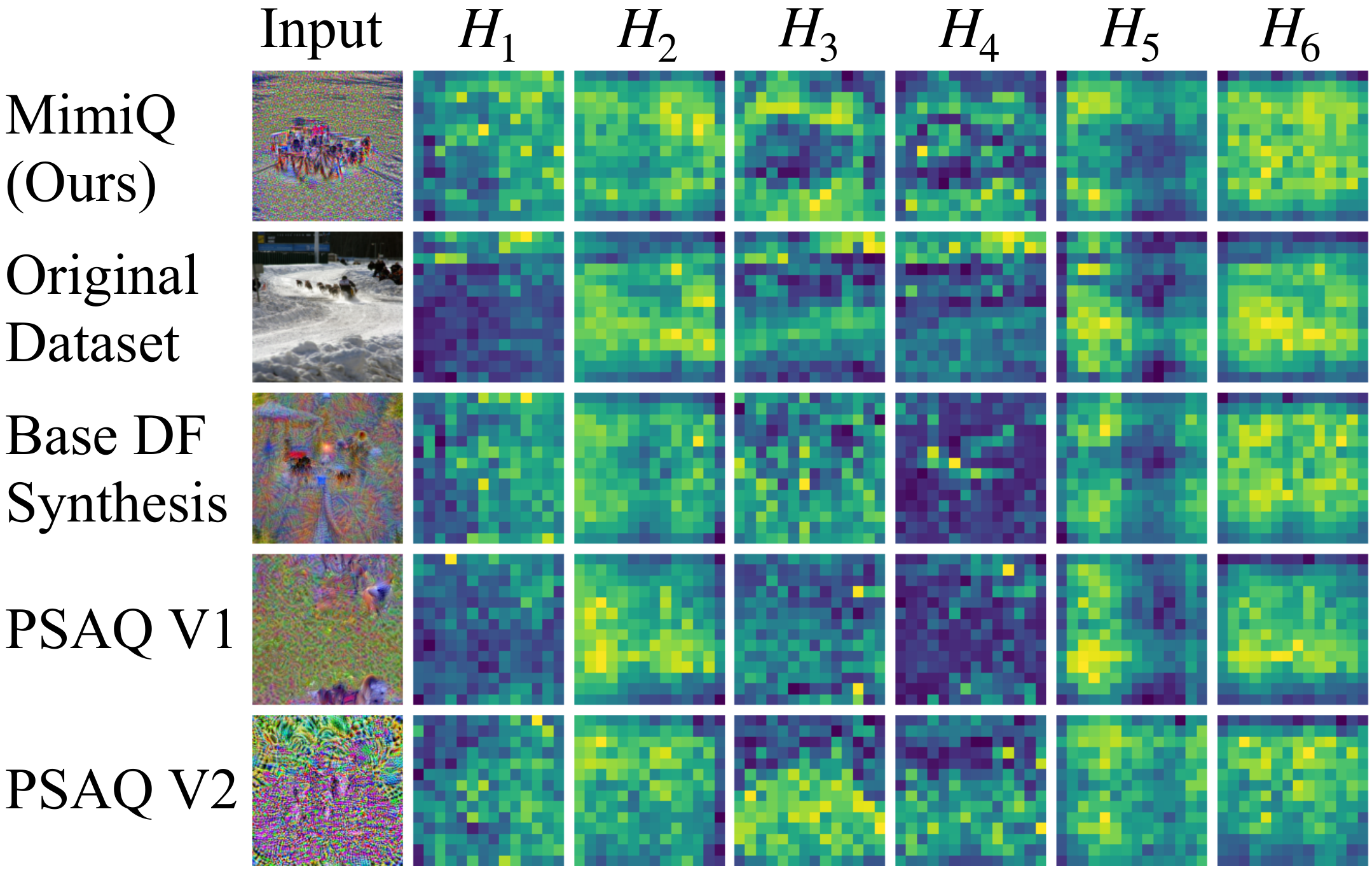}
\vspace{3mm}
\end{subfigure}
\begin{subfigure}[t]{0.6\textwidth}
\centering
\includegraphics[width=0.93\textwidth]{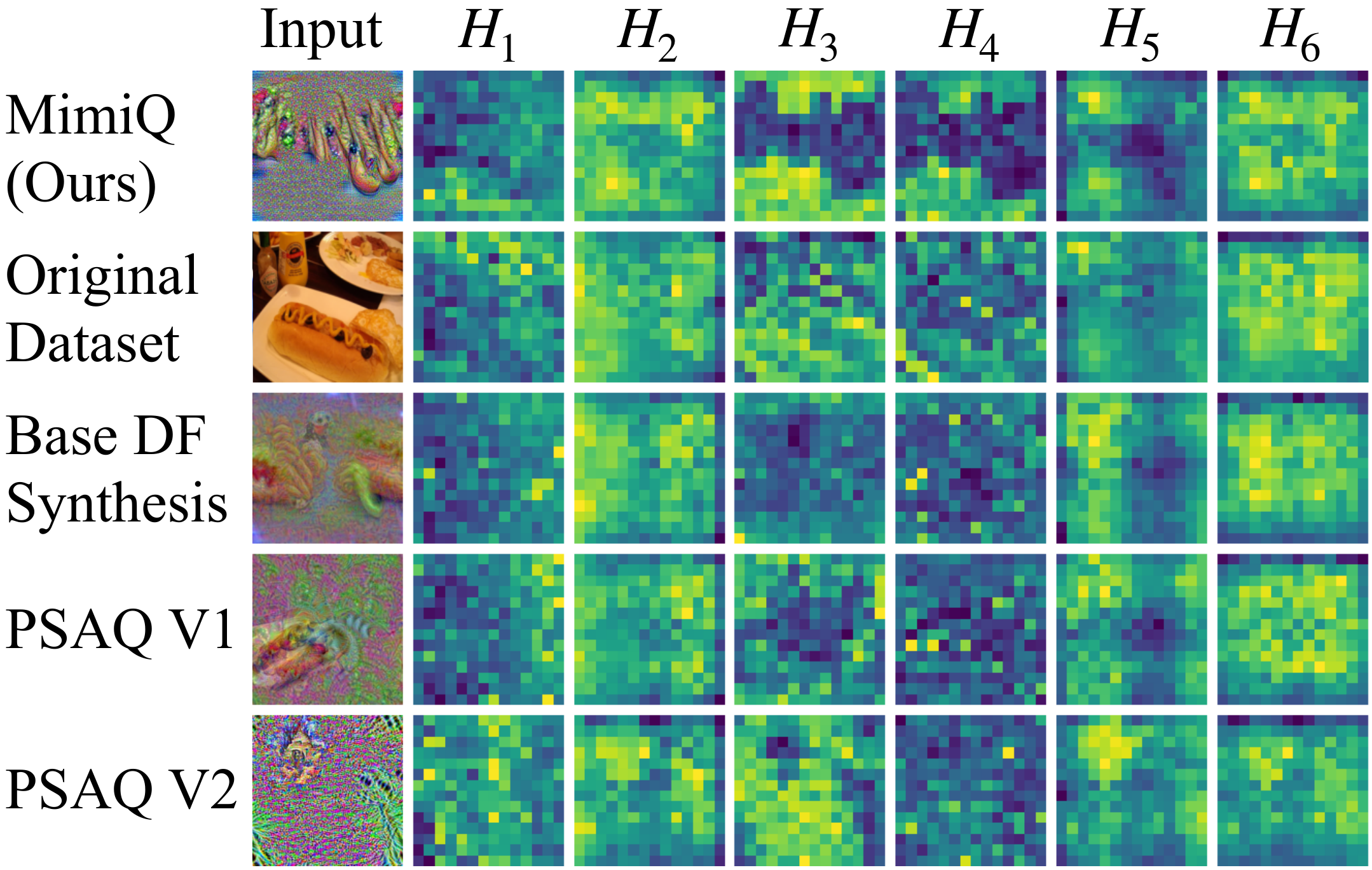}
\end{subfigure}

\caption{Further attention map visualization of the input image from \aname, Real training samples, and DFQ baselines. The attention map is generated from \textbf{DeiT-Small} architecture. Each row represents different input images and each column represents one of the attention heads. }
\label{fig:synth_compare_ds}
\end{figure*}

\begin{figure*}[t]
\centering
\begin{subfigure}[t]{0.3\textwidth}
\centering
\includegraphics[width=\textwidth]{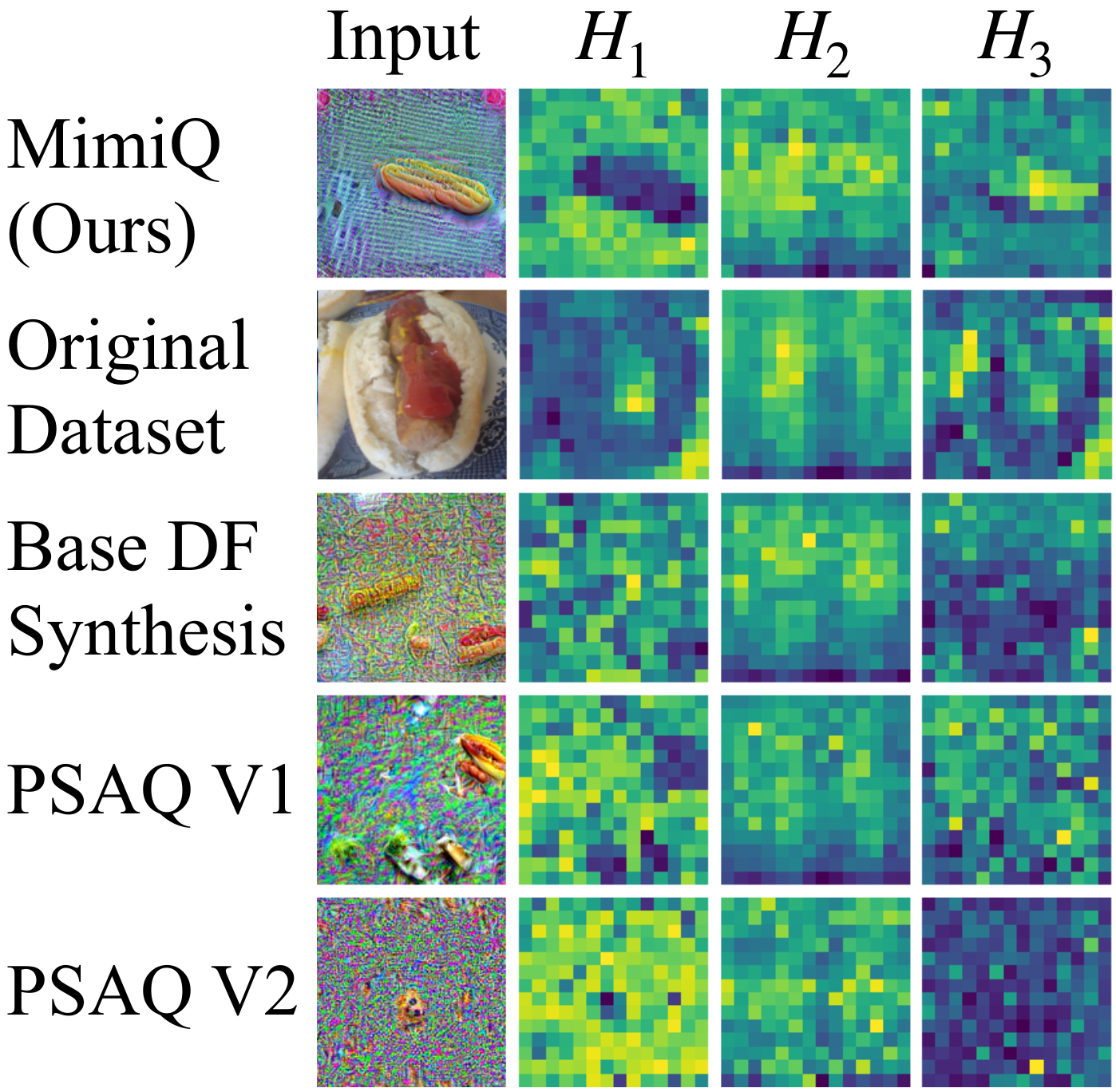}
\end{subfigure}
\begin{subfigure}[t]{0.3\textwidth}
\centering
\includegraphics[width=\textwidth]{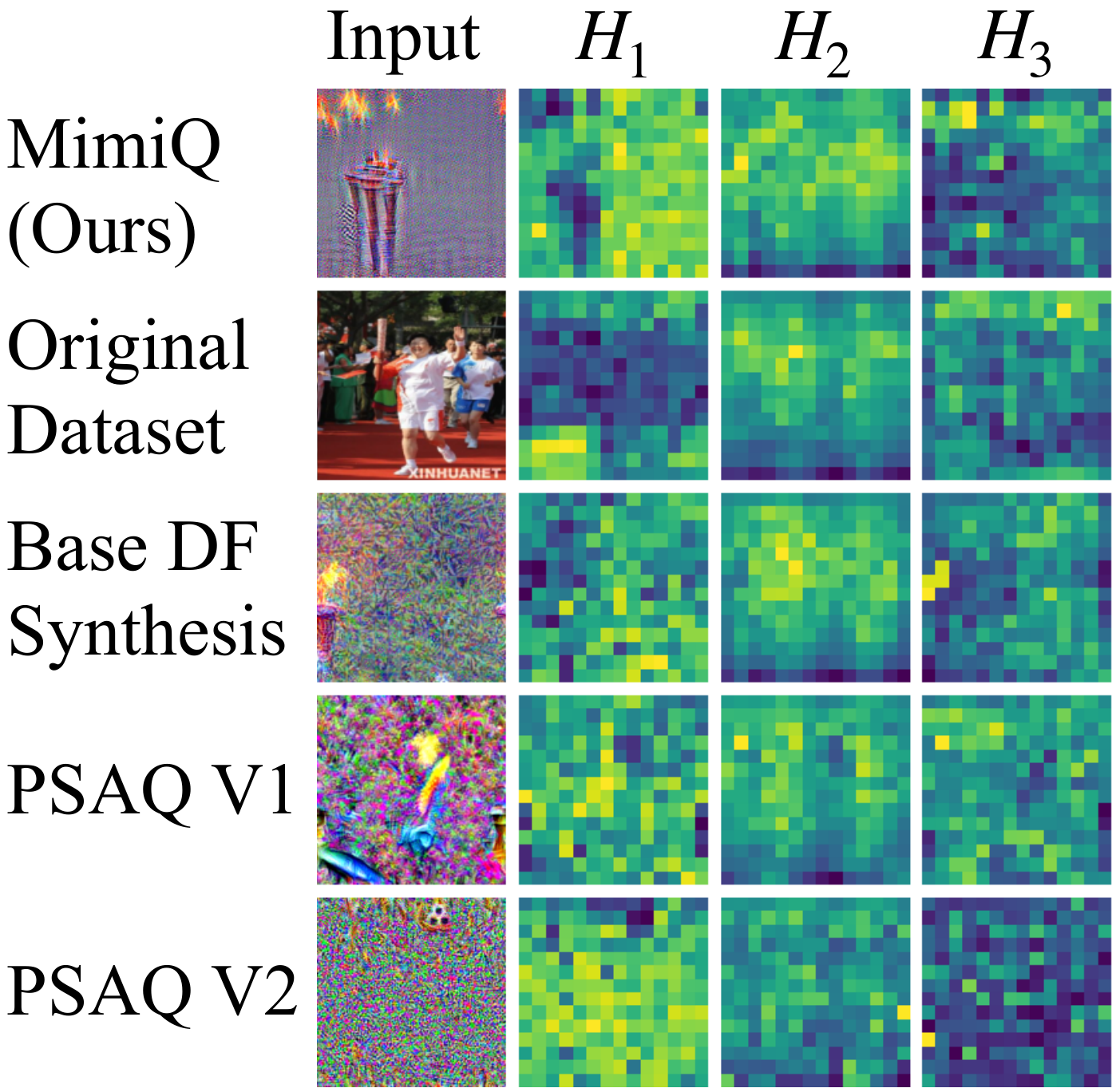}
\end{subfigure}
\begin{subfigure}[t]{0.3\textwidth}
\centering
\includegraphics[width=\textwidth]{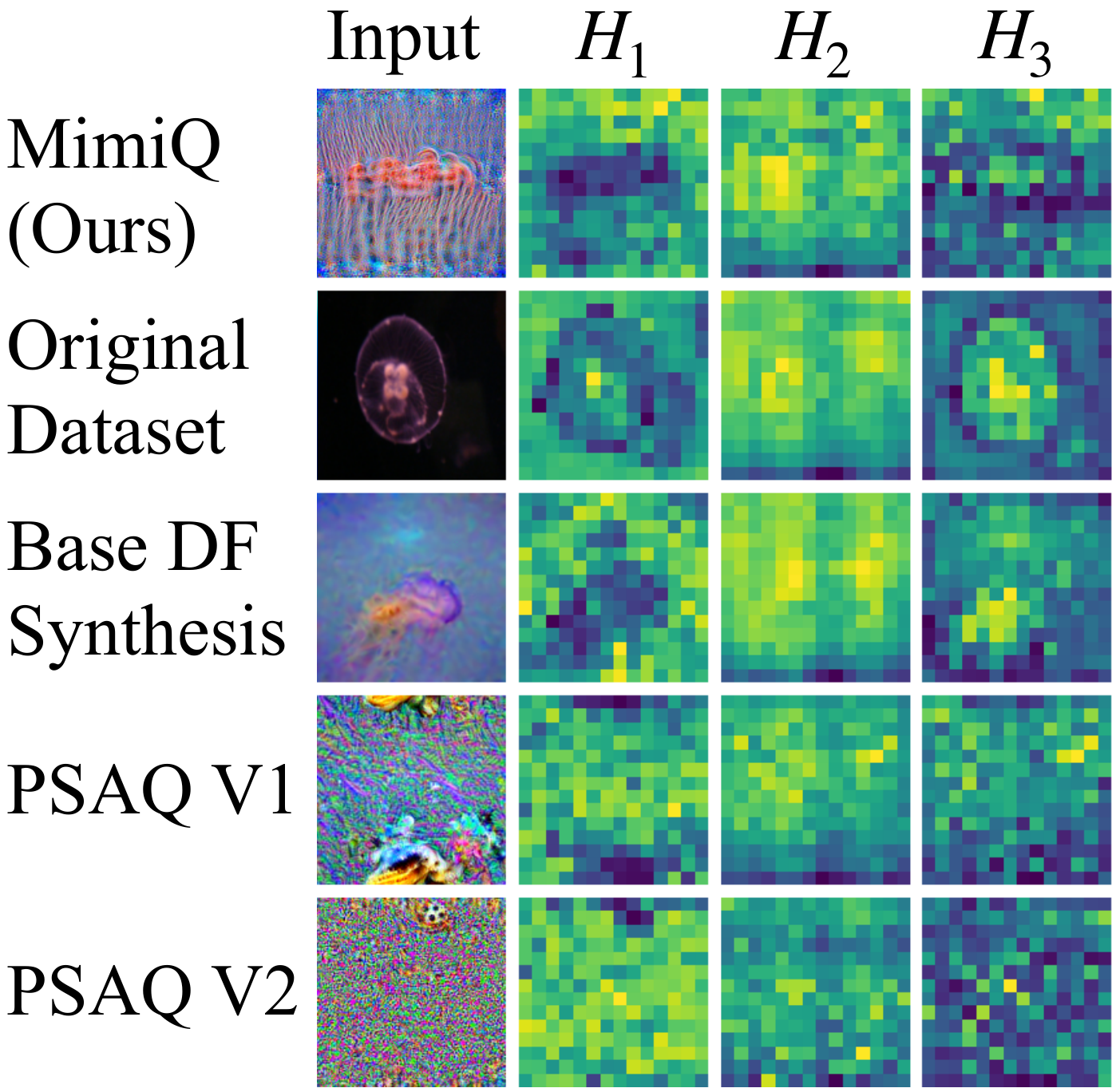}
\end{subfigure}

\caption{Further attention map visualization of the input image from \aname, Real training samples, and DFQ baselines. The attention map is generated from \textbf{ViT-Tiny} architecture. Each row represents different input images and each column represents one of the attention heads. }
\label{fig:synth_compare_vt}
\end{figure*}

\begin{figure*}[t]
\centering
\begin{subfigure}[t]{0.3\textwidth}
\centering
\includegraphics[width=\textwidth]{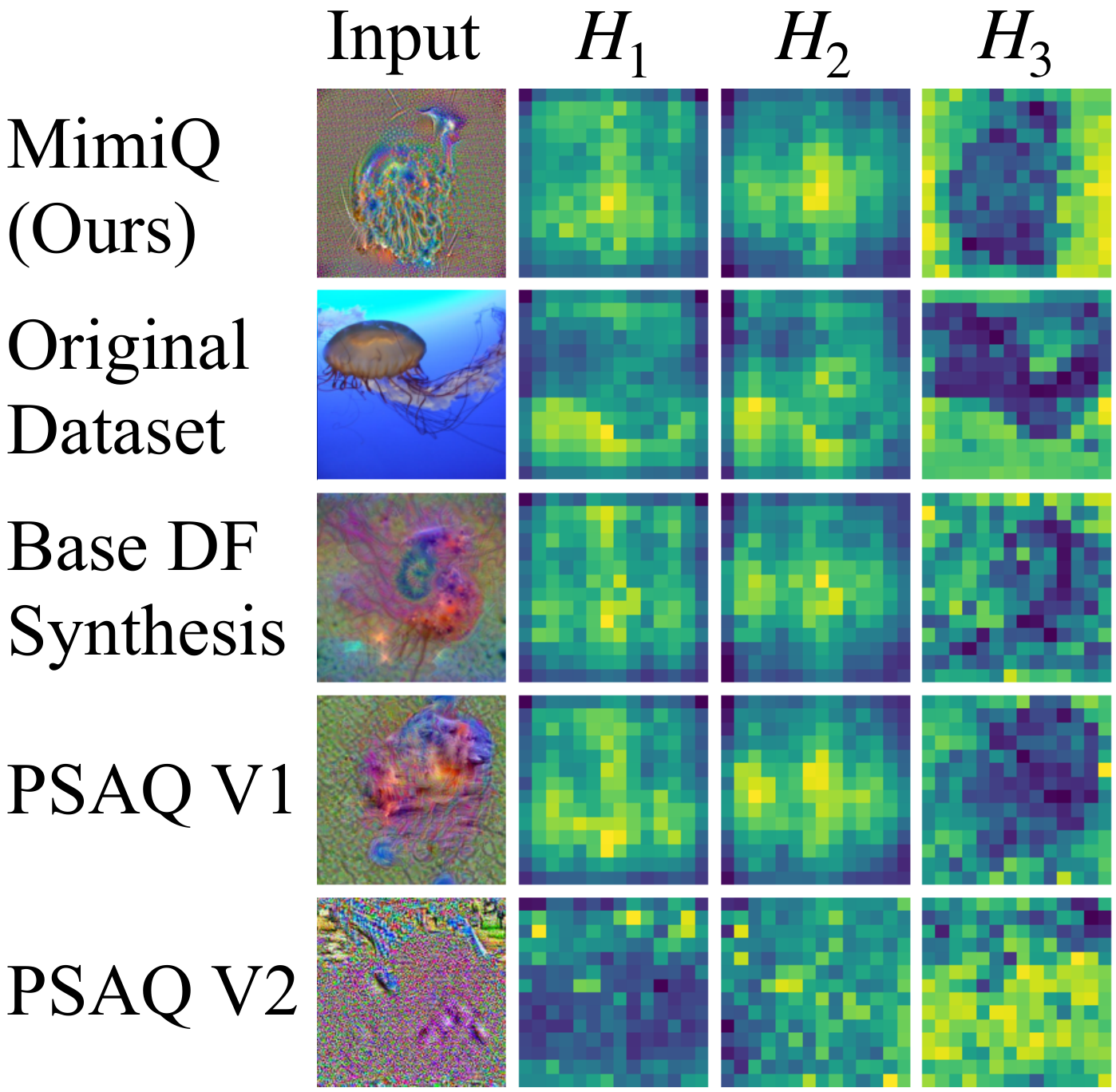}
\end{subfigure}
\begin{subfigure}[t]{0.3\textwidth}
\centering
\includegraphics[width=\textwidth]{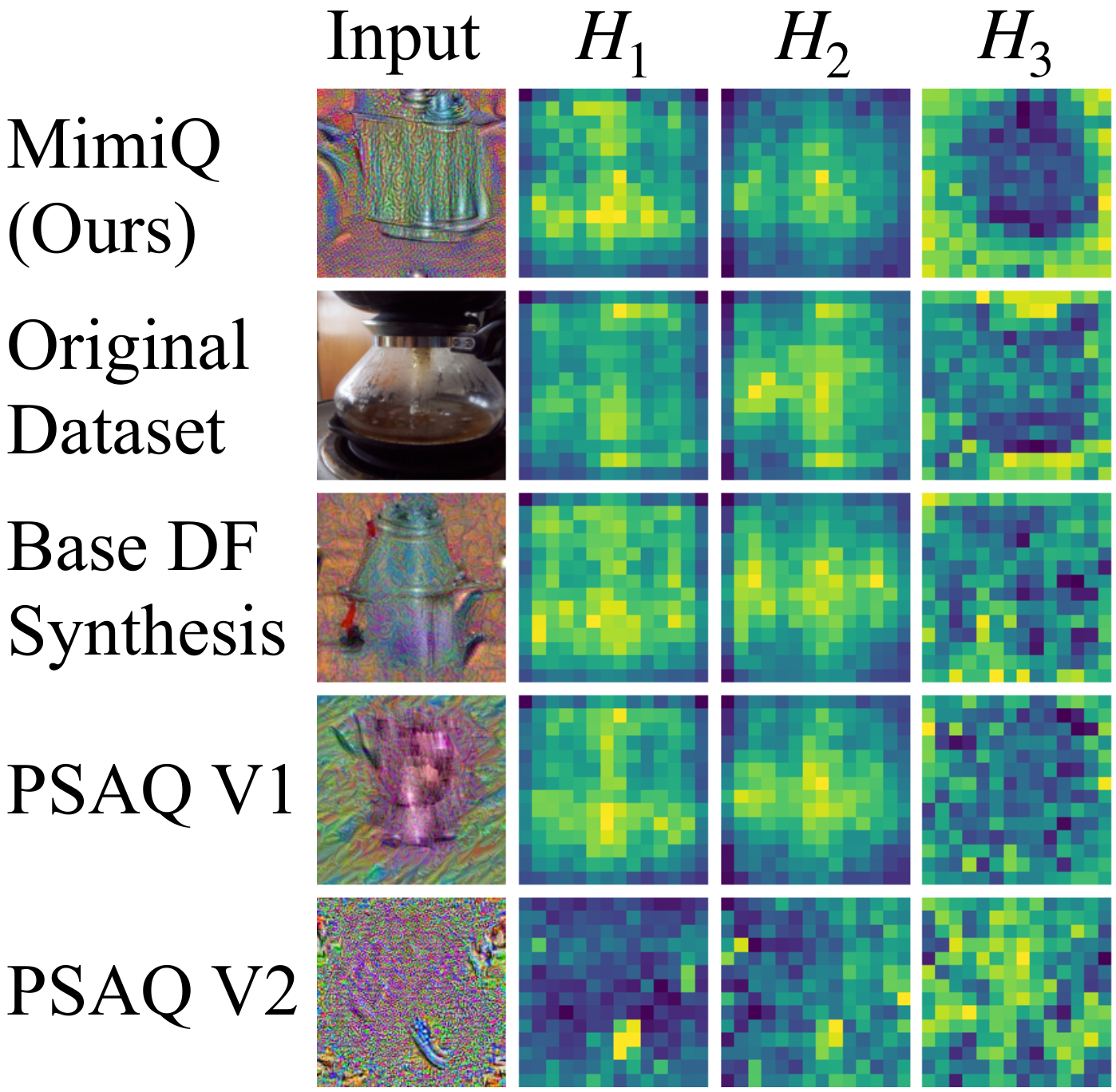}
\end{subfigure}
\begin{subfigure}[t]{0.3\textwidth}
\centering
\includegraphics[width=\textwidth]{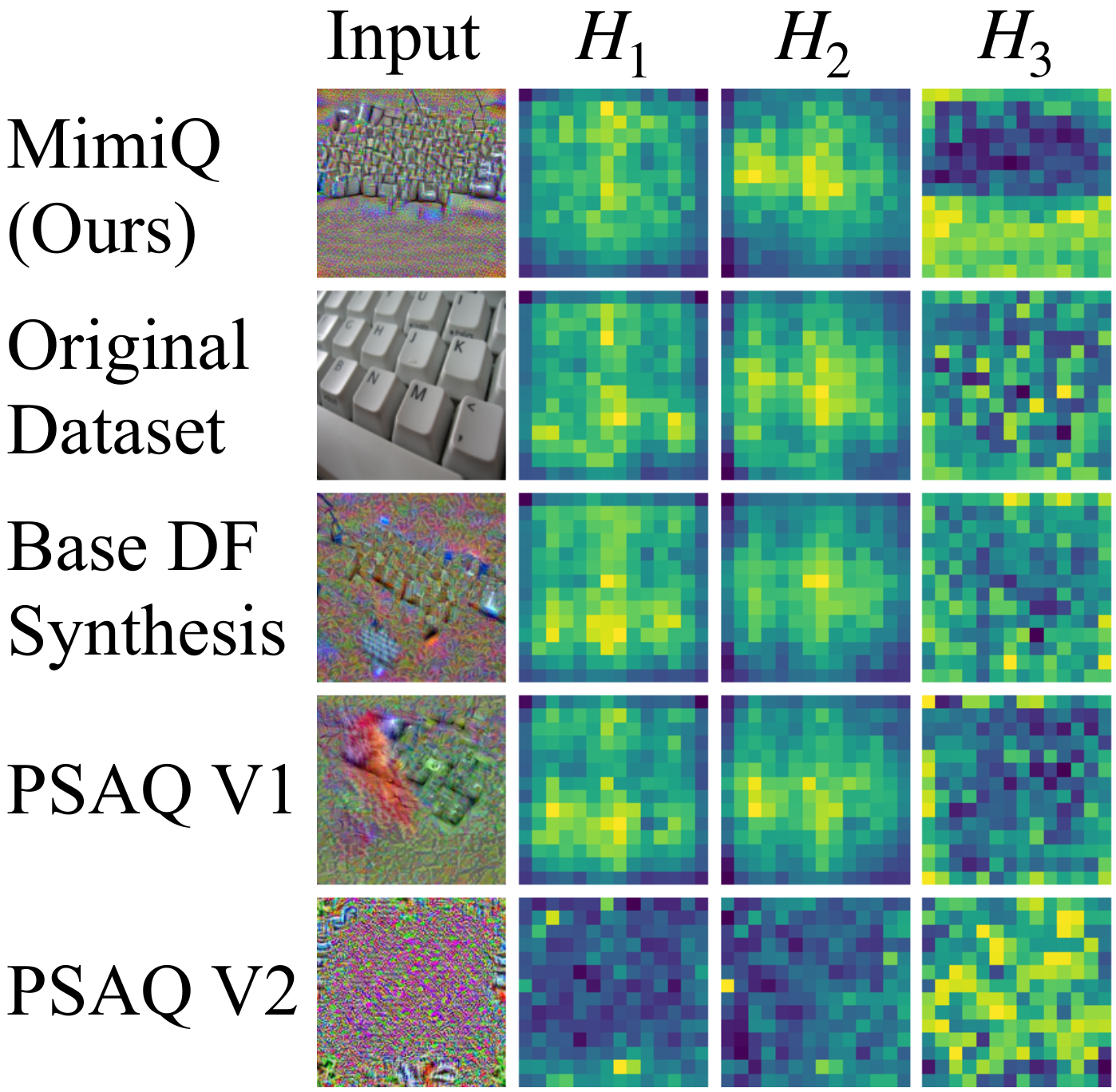}
\end{subfigure}

\caption{Further attention map visualization of the input image from \aname, Real training samples, and DFQ baselines. The attention map is generated from \textbf{DeiT-Tiny} architecture. Each row represents different input images and each column represents one of the attention heads. }
\label{fig:synth_compare_dt}
\end{figure*}

In \cref{fig:motiv_map} of the main body, we visualize the attention map of synthetic and real samples using ViT-Base architecture.
Here, we plot more attention map visualization from various network architectures with baseline synthetic dataset generation methods in \cref{fig:synth_compare_vb,fig:synth_compare_vs,fig:synth_compare_vt,fig:synth_compare_db,fig:synth_compare_ds,fig:synth_compare_dt}.
In all target network architectures, \aname shows coherent attention map structures among multiple attention heads, which are close to that from the real samples. 
In -Small and -Tiny networks, which have fewer attention heads, i.e., three for the -Tiny and six for the -Small, the attention map from the baseline reconstruction methods did not get a clear attention structure.
However, despite the smaller number of attention heads, \aname still preserves better attention similarity among attention heads.

\section{Comparison of Batch Normalization and Layer Normalization}

Here, we provide a detailed explanation of what BN-based dataset reconstruction loss is and why it cannot be used for ViT architectures.
The BN layer applies normalization and linear transformation to input activation $x$ with batch dimension $\mathcal{B}$ and feature dimension $K$ as follows:
\begin{align}
    BN(x) &= \gamma_{bn} \frac{x_k - \mu_k}{\sqrt{\sigma^2_k + \epsilon}} + \beta_{bn}, \\
     \mu_k&=\frac{1}{\mathcal{B}} \sum^\mathcal{B}_{j=1} x_{j,k}, \\
    \sigma^2_k &= \frac{1}{\mathcal{B}} \sum^\mathcal{B}_{j=1} (x_{j,k} - \mu_k)^2,
\end{align}
where $x_{j,k}$ is $k$-th feature dimension of $j$-th input batch from input $x$, $\mu_k$ is feature mean, $\sigma^2_k$ is feature variance, $\gamma_{bn}$ and $\beta_{bn}$ are trainable. %
In the training phase, $\mu_k$ and $\sigma_k$ store feature statistics of the training dataset and use stored statistics in the inference phase. 

By utilizing these feature statistics, current DFQ methods targeting BN-based CNN generate synthetic datasets closer to the original dataset. 
To achieve this, synthetic datasets are generated minimizing batch-norm statistics loss ($\mathcal{L}_{BNS}$): 
\begin{align}
    \mathcal{L}_{BNS}(\hat x) &= \sum^L_l \sum^K_k \lVert \hat \mu_{l,k} - \mu_{l,k} \lVert^2_2 + \lVert \hat \sigma_{l,k} - \sigma_{l,k} \lVert^2_2,
\end{align}
where $L$ is the number of BN layers in the target network, $\hat \mu$, $\hat \sigma$ are feature statistics from a synthetic sample.
As a result, synthetic datasets follow similar statistical characteristics to the original training dataset at a feature level. 

However, the layer normalization layer (LN) does not store the statistical properties of the original training dataset. 
Similar to BN, LN applies normalization and linear transformation to the input features as:
\begin{align}
    LN(x) &= \gamma_{ln} \frac{x - \mu_x}{\sqrt{\sigma^2_x + \epsilon}} + \beta_{ln}, \\
     \mu_x&=\frac{1}{K} \sum^K_{k=1} x_k, \\
    \sigma^2_x &= \frac{1}{K} \sum^K_{k=1} (x_k - \mu_x)^2,
\end{align}
where mean and variance are calculated from input $x$. 
Therefore, LN does not utilize saved feature statistics but calculates mean ($\mu_x$) and variance ($\sigma^2_x$) directly from input.
As ViT utilizes LN as a normalization layer, existing BN-based DFQ methods cannot use $\mathcal{L}_{BNS}$, which is crucial to generate a synthetic dataset closer to the original training data.

\begin{table}[b]
    \centering

    \resizebox{\columnwidth}{!}
    {
    \begin{tabular}{clcccc}
    \toprule
    \multirow{2}{*}{Bits} & \multirow{2}{*}{Methods}  &  \multicolumn{4}{c}{Networks} \\
    \cmidrule{3-6}  
       & & ViT-B & DeiT-T & DeiT-S  & DeiT-B  \\
    \midrule
\multirow{2}{*}{W4/A4}& CLAMP-ViT & 		0.10	&	0.15	&	0.13	&	0.11	\\ 
&{\aname (Ours)}  & 		62.91	&	52.03	&	62.72	&	74.10	\\ \midrule
\multirow{2}{*}{W5/A5}& CLAMP-ViT & 		0.08	&	5.42	&	3.19	&	0.17	\\
&{\aname (Ours)}  & 		78.09	&	63.40	&	72.59	&	78.20	\\ \midrule
\multirow{2}{*}{W4/A8}& CLAMP-ViT & 		81.42	&	65.71	&	76.44	&	79.76	\\
&{\aname (Ours)}  & 		84.20	&	69.86	&	78.48	&	81.34	\\ \midrule
\multirow{2}{*}{W8/A8}& CLAMP-ViT & 		82.55	&	71.76	&	79.56	&	81.44	\\
&{\aname (Ours)}  & 		85.17	&	72.01	&	79.73	&	81.87	\\ 

    \bottomrule
    \end{tabular}
    }
   
        \caption{ImageNet classification accuracy comparison with ~\citet{ramachandran2024clamp}.}
    \label{tab:supp:clamp}
\end{table}

\section{Comparison with CLAMP-ViT}

Here, we compare image classification accuracy with CLAMP-ViT~\cite{ramachandran2024clamp}, which the preprint was made available online concurrently with our work. 
As mentioned in the main body of the paper, the direct comparison is inadequate due to the differences in the experimental setups, including quantization settings, pre-trained weights, etc. 
To ensure a fair comparison, we faithfully implemented CLAMP-ViT following the original paper and the evaluation code provided by the authors, applying our best efforts to match the conditions as closely as possible. 
The comparison is shown in \cref{tab:supp:clamp}.
The experimental results indicate that CLAMP-ViT significantly underperforms under the low-bit quantization settings, which suggests that CLAMP-ViT shares the same limitation observed in other baselines. 
This further supports our observation that relying solely on patch-wise similarity is insufficient to fully capture the embedded information from the pretrained ViT network, which is related to the inter-head attention map dissimilarity problem (\cref{fig:motiv_map}).

\section{Measuring Inter-head Similarity with SSIM}
\label{sec:ssim_explain}

When measuring the similarity between different heads, it is essential to consider both positive and negative correlations. 
The real-world images show that attention maps from one head often produce similar or inverted outputs to the other heads (as shown in \cref{fig:motiv_map}). 
Therefore, using metrics that only detect positive correlation, such as the mean-square error (MSE), is insufficient.

To capture both positive and negative correlations, we utilize the structural similarity index measure (SSIM)~\cite{ssim} when measuring head-wise attention map similarity.
SSIM compares two images ($I_x,I_y$) from three perspectives: luminance ($l(I_x,I_y)$), contrast ($c(I_x,I_y)$), and structure ($s(I_x,I_y)$), utilizing mean ($\mu$), standard deviation ($\sigma$), and covariance ($\sigma_{xy}$), respectively. 
Thus, each component can be calculated as follows:
\begin{align}
    l(I_x,I_y) &= \frac{2\mu_x\mu_y+c_1}{\mu^2_x + \mu^2_y +c_1}, \label{eq:ssim_lum}\\
    c(I_x,I_y) &= \frac{2\sigma_x\sigma_y+c_2}{\sigma^2_x + \sigma^2_y +c_2}, \label{eq:ssim_con}\\
    s(I_x,I_y) &= \frac{\sigma_{xy} + c_3}{\sigma_x \sigma_y + c_3}. \label{eq:ssim_str}
\end{align}
SSIM is calculated using the weighted product of luminance ($l$), contrast ($c$), and structural similarity ($s$):
\begin{align}
    SSIM(I_x,I_y) &= l(I_x,I_y)^\alpha \cdot c(I_x,I_y)^\beta \cdot s(I_x,I_y)^\gamma, \notag  \\
    &= \frac{(2\mu_x\mu_y+c_1)(2\sigma_{xy} + c_2)}{(\mu^2_x + \mu^2_y +c_1)(\sigma^2_x + \sigma^2_y +c_2)},
    \label{eq:ssim}
\end{align}
where, $\alpha=\beta=\gamma=1$ and $c_3=c_2/2$.
One crucial aspect of SSIM is that it not only measures positive correlation ($SSIM>0.0$), but also negative correlation ($SSIM<0.0$), which is necessary when measuring inter-head similarity. 
A positive SSIM value means that the two input images are similar, while a negative SSIM value implies that one input image is close to an inverted version of the other. 
Therefore, we use SSIM to measure the inter-head similarity of the MSA layer.

\section{Extended Experimental Results of Image Classification Task}

We present the extended experimental results of an image classification task, including the quantization accuracies for each min-max and LSQ activation quantization setting, as well as additional results from the W6/A6 quantization setting. 

The results of the min-max activation quantization are presented in \cref{tab:minmax_full}. 
Overall, our proposed method, \aname, outperforms both CNN-targeting baselines and ViT-aware baselines. 
Compared to LSQ activation quantization, min-max quantization generally exhibits higher quantization accuracy in higher-bit settings while showing opposite trends in lower-bit settings. 
We attribute this to the simplicity of the min-max quantization method, which only relies on the activation's minimum and maximum values. 
Quantized parameters with higher-bit settings can preserve more information by allowing a finer quantization grid, making simple min-max quantization more robust to distributional outliers with significantly high magnitudes. 
However, in low-bit settings, the outliers may harm the quantization accuracy as quantization becomes more susceptible to outliers due to lower representational capacity. 
Nonetheless, \aname consistently outperforms the baselines by a significant margin, demonstrating the effectiveness of our proposed method.

We also report the LSQ activation quantization results in \cref{tab:lsq_full}. 
The experimental results show that LSQ significantly boosts quantization accuracy in low-bit settings. 
This is due to the characteristic of LSQ, which determines the quantization scale to reduce task losses. 
Therefore, the quantized parameter may be more robust to the min-max distribution, which is easily affected by outliers. 
As LSQ requires gradient-based fine-tuning of quantization parameters ($s$,$z$ in \cref{eq:q_basic}), the computation of the min-max and LSQ quantization in the inference phase is identical. 
Thus, they have the same memory consumption and inference latency.

\begin{table*}[]
    \centering

     \def\arraystretch{0.85}%
    \resizebox{0.92\textwidth}{!}
    {
    \begin{tabular}{clcccccccccc}
    \toprule
    \multirow{2}{*}{Bits} & \multirow{2}{*}{Methods} & \multirow{2}{*}{\makecell{Target \\ Arch.}} &  &  & && Models &&& & \\
    \cmidrule{4-12}  
       & &  &  ViT-T & ViT-S  & ViT-B & DeiT-T & DeiT-S  & DeiT-B & Swin-T & Swin-S  & Swin-B \\
    \midrule
\multirow{8}{*}{W4/A4}& Real-Data FT & - &		27.79	&	54.20	&	67.81	&	51.01	&	44.10	&	46.73	&	67.69	&	76.34	&	62.83	\\ \cmidrule{2-12}						
& GDFQ & CNN&		\textcolor{white}00.52	&	\textcolor{white}02.74	&	\textcolor{white}01.03	&	\textcolor{white}04.84	&	\textcolor{white}06.81	&	\textcolor{white}08.70	&	\textcolor{white}04.25	&	41.93	&	\textcolor{white}00.30	\\ 						
& Qimera & CNN&		\textcolor{white}00.52	&	\textcolor{white}07.02	&	\textcolor{white}05.61	&	15.18	&	11.37	&	32.49	&	11.28	&	39.64	&	29.27	\\ 						
& AdaDFQ & CNN&		\textcolor{white}00.56	&	\textcolor{white}01.78	&	\textcolor{white}00.21	&	\textcolor{white}09.79	&	\textcolor{white}04.64	&	\textcolor{white}02.91	&	\textcolor{white}00.43	&	39.40	&	\textcolor{white}01.55	\\						
&PSAQ-ViT V1 & ViT &		\textcolor{white}00.00	&	\textcolor{white}00.00	&	\textcolor{white}00.00	&	\textcolor{white}01.70	&	\textcolor{white}01.45	&	\textcolor{white}01.19	&	\textcolor{white}00.10	&	\textcolor{white}01.19	&	\textcolor{white}00.00	\\ 						
& PSAQ-ViT V2 & ViT &		\textcolor{white}00.35	&	\textcolor{white}00.92	&	\textcolor{white}00.19	&	\textcolor{white}06.00	&	\textcolor{white}04.66	&	\textcolor{white}03.59	&	\textcolor{white}01.70	&	39.10	&	20.36	\\ \cmidrule{2-12}						
& \multirow{2}{*}{\aname (Ours)}  & \multirow{2}{*}{ViT} &		13.02	&	32.68	&	49.22	&	33.61	&	24.40	&	51.77	&	58.56	&	70.46	&	66.95	\\						
&&&		+12.46	&	+25.66	&	+43.61	&	+18.44	&	+13.03	&	+19.27	&	+47.28	&	+28.53	&	+37.68	\\ \midrule						
																									
\multirow{8}{*}{W5/A5}& Real-Data FT & - &		59.82	&	72.85	&	80.52	&	65.23	&	73.95	&	78.39	&	78.71	&	81.74	&	83.08	\\ \cmidrule{2-12}						
& GDFQ & CNN&		24.40	&	53.96	&	29.18	&	44.76	&	57.00	&	71.03	&	42.29	&	78.04	&	70.55	\\ 						
& Qimera & CNN&		26.70	&	16.13	&	\textcolor{white}09.43	&	33.13	&	33.65	&	47.01	&	62.13	&	46.81	&	43.57	\\ 						
& AdaDFQ & CNN&		27.10	&	59.36	&	43.02	&	53.85	&	59.55	&	71.12	&	33.42	&	79.82	&	75.59	\\ 						
&PSAQ-ViT V1 & ViT &		\textcolor{white}03.39	&	29.36	&	\textcolor{white}06.12	&	41.97	&	47.35	&	57.23	&	10.38	&	76.33	&	67.75	\\ 						
& PSAQ-ViT V2 & ViT &		23.05	&	63.59	&	74.29	&	55.18	&	65.30	&	73.16	&	53.49	&	80.55	&	79.80	\\ \cmidrule{2-12}						
& \multirow{2}{*}{\aname (Ours)}  & \multirow{2}{*}{ViT} &		43.21	&	66.92	&	78.09	&	59.99	&	69.00	&	76.34	&	76.39	&	80.75	&	82.05	\\						
&&&		+16.11	&	+3.33	&	+3.80	&	+4.81	&	+3.70	&	+3.18	&	+14.27	&	+0.20	&	+2.25	\\ \midrule						
																									
\multirow{8}{*}{W6/A6}& Real-Data FT & - &		70.64	&	79.07	&	83.96	&	70.16	&	78.30	&	81.11	&	80.45	&	82.73	&	84.52	\\ \cmidrule{2-12}						
& GDFQ & CNN&		64.10	&	75.73	&	69.98	&	65.38	&	74.29	&	79.09	&	78.70	&	82.48	&	83.81	\\ 						
& Qimera & CNN&		66.09	&	36.83	&	29.72	&	55.04	&	55.80	&	60.93	&	71.18	&	62.56	&	64.17	\\ 						
& AdaDFQ & CNN&		62.63	&	73.21	&	73.75	&	68.30	&	74.60	&	80.44	&	79.36	&	82.60	&	84.04	\\ 						
&PSAQ-ViT V1 & ViT &		53.23	&	69.20	&	79.69	&	68.11	&	74.07	&	80.04	&	79.06	&	82.67	&	84.41	\\ 						
& PSAQ-ViT V2 & ViT &		65.40	&	77.13	&	83.42	&	69.08	&	77.46	&	80.94	&	79.69	&	82.66	&	84.16	\\ \cmidrule{2-12}						
& \multirow{2}{*}{\aname (Ours)}  & \multirow{2}{*}{ViT} &		66.28	&	77.81	&	83.71	&	69.24	&	77.77	&	81.15	&	80.16	&	82.60	&	84.08	\\						
&&&		+0.19	&	+0.68	&	+0.28	&	+0.16	&	+0.31	&	+0.21	&	+0.47	&	-0.07	&	-0.33	\\ \midrule						
																									
\multirow{8}{*}{W4/A8}& Real-Data FT & - &		71.27	&	79.84	&	84.52	&	70.37	&	78.93	&	81.47	&	80.47	&	82.46	&	84.29	\\ \cmidrule{2-12}						
& GDFQ & CNN&		62.65	&	76.06	&	81.68	&	65.82	&	76.49	&	80.03	&	78.90	&	81.47	&	83.63	\\ 						
& Qimera & CNN&		61.80	&	60.08	&	63.22	&	61.90	&	70.10	&	72.38	&	73.93	&	72.22	&	76.35	\\ 						
& AdaDFQ & CNN&		64.67	&	76.27	&	82.43	&	67.71	&	76.92	&	80.49	&	79.70	&	82.07	&	83.78	\\ 						
&PSAQ-ViT V1& ViT &		59.95	&	75.93	&	82.01	&	66.16	&	76.56	&	80.05	&	79.06	&	81.89	&	84.07	\\ 						
& PSAQ-ViT V2 & ViT &		66.68	&	78.24	&	84.02	&	68.23	&	78.27	&	81.15	&	79.98	&	82.04	&	83.90	\\ \cmidrule{2-12}						
& \multirow{2}{*}{\aname (Ours)}  & \multirow{2}{*}{ViT} &		68.11	&	78.77	&	84.20	&	69.86	&	78.48	&	81.34	&	80.06	&	82.08	&	83.09	\\						
&&&		+1.43	&	+0.53	&	+0.18	&	+1.63	&	+0.21	&	+0.20	&	+0.08	&	+0.01	&	+0.99	\\ \midrule						
																									
\multirow{8}{*}{W8/A8}& Real-Data FT & - &		74.83	&	81.30	&	85.13	&	71.99	&	79.70	&	81.77	&	80.96	&	83.08	&	84.79	\\ \cmidrule{2-12}						
& GDFQ & CNN&		72.90	&	80.97	&	84.81	&	71.83	&	79.59	&	81.62	&	80.83	&	82.99	&	84.42	\\ 						
& Qimera & CNN&		72.88	&	81.04	&	84.98	&	71.76	&	79.46	&	81.58	&	80.41	&	82.95	&	84.37	\\ 						
& AdaDFQ & CNN&		73.84	&	81.11	&	84.88	&	71.72	&	79.34	&	81.73	&	80.89	&	82.99	&	84.70	\\ 						
&PSAQ-ViT V1& ViT &		72.73	&	81.17	&	84.89	&	71.99	&	79.71	&	81.79	&	81.26	&	83.29	&	85.13	\\ 						
& PSAQ-ViT V2 & ViT &		73.43	&	81.25	&	85.11	&	71.90	&	79.70	&	81.86	&	80.88	&	83.00	&	84.71	\\ \cmidrule{2-12}						
& \multirow{2}{*}{\aname (Ours)}  & \multirow{2}{*}{ViT} &		74.60	&	81.30	&	85.17	&	72.01	&	79.73	&	81.87	&	80.96	&	83.05	&	84.79	\\						
&&&		+0.76	&	+0.05	&	+0.07	&	+0.02	&	+0.02	&	+0.01	&	-0.30	&	-0.24	&	-0.34	\\ \bottomrule						

    \end{tabular}}
        \caption{Experimental results on ImageNet image classification dataset with min-max activation quantization.}
    \label{tab:minmax_full}
\end{table*}

\begin{table*}[]
    \centering

     \def\arraystretch{0.85}%
    \resizebox{0.92\textwidth}{!}
    {
    \begin{tabular}{clcccccccccc}
    \toprule
    \multirow{2}{*}{Bits} & \multirow{2}{*}{Methods} & \multirow{2}{*}{\makecell{Target \\ Arch.}} &  &  & && Models &&& & \\
    \cmidrule{4-12}  
       & &  &  ViT-T & ViT-S  & ViT-B & DeiT-T & DeiT-S  & DeiT-B & Swin-T & Swin-S  & Swin-B \\
    \midrule
\multirow{8}{*}{W4/A4}& Real-Data FT & - &		58.17	&	67.21	&	57.42	&	57.98	&	62.15	&	64.96	&	73.08	&	68.33	&	73.06	\\ \cmidrule{2-12}						
& GDFQ & CNN&		\textcolor{white}02.95	&	\textcolor{white}04.62	&	11.73	&	25.96	&	22.12	&	30.04	&	42.08	&	35.11	&	36.04	\\ 						
& Qimera & CNN&		\textcolor{white}00.57	&	\textcolor{white}01.59	&	\textcolor{white}01.62	&	11.33	&	6.94	&	23.16	&	47.98	&	28.78	&	24.19	\\ 						
& AdaDFQ & CNN&		\textcolor{white}02.00	&	\textcolor{white}01.37	&	\textcolor{white}06.21	&	19.57	&	14.44	&	19.22	&	38.88	&	16.19	&	32.26	\\						
&PSAQ-ViT V1 & ViT &		\textcolor{white}00.67	&	\textcolor{white}00.15	&	\textcolor{white}00.94	&	19.61	&	\textcolor{white}05.90	&	\textcolor{white}08.74	&	22.71	&	\textcolor{white}09.26	&	23.69	\\ 						
& PSAQ-ViT V2 & ViT &		\textcolor{white}01.54	&	\textcolor{white}04.14	&	\textcolor{white}02.83	&	22.82	&	32.57	&	45.81	&	50.42	&	25.77	&	39.26	\\ \cmidrule{2-12}						
& \multirow{2}{*}{\aname (Ours)}  & \multirow{2}{*}{ViT} &		42.99	&	55.69	&	62.91	&	52.03	&	62.72	&	74.10	&	69.33	&	66.36	&	73.49	\\						
&&&		+40.04	&	+51.07	&	+51.18	&	+26.07	&	+30.15	&	+28.28	&	+18.91	&	+31.25	&	+34.23	\\ \midrule						
																									
\multirow{8}{*}{W5/A5}& Real-Data FT & - &		68.49	&	73.90	&	73.17	&	66.10	&	72.31	&	73.60	&	77.21	&	75.24	&	78.00	\\ \cmidrule{2-12}						
& GDFQ & CNN&		10.24	&	24.21	&	33.56	&	43.84	&	49.16	&	49.59	&	61.30	&	54.94	&	54.71	\\ 						
& Qimera & CNN&		\textcolor{white}03.61	&	16.13	&	\textcolor{white}09.43	&	33.13	&	33.65	&	47.01	&	62.13	&	46.81	&	43.57	\\ 						
& AdaDFQ & CNN&		15.10	&	20.01	&	27.14	&	42.78	&	42.15	&	41.76	&	64.61	&	51.43	&	57.64	\\ 						
&PSAQ-ViT V1 & ViT &		17.66	&	23.37	&	16.80	&	53.36	&	29.37	&	42.38	&	58.63	&	43.85	&	57.80	\\ 						
& PSAQ-ViT V2 & ViT &		40.21	&	49.87	&	57.26	&	54.49	&	63.95	&	67.05	&	69.77	&	59.88	&	66.14	\\ \cmidrule{2-12}						
& \multirow{2}{*}{\aname (Ours)}  & \multirow{2}{*}{ViT} &		62.40	&	70.02	&	77.87	&	63.40	&	72.59	&	78.20	&	75.66	&	74.58	&	78.62	\\						
&&&		+22.19	&	+20.15	&	+20.60	&	+8.91	&	+8.63	&	+11.15	&	+5.90	&	+14.70	&	+12.48	\\ \midrule						
																									
\multirow{8}{*}{W6/A6}& Real-Data FT & - &		72.44	&	77.43	&	79.86	&	69.35	&	76.07	&	77.44	&	78.89	&	78.60	&	81.34	\\ \cmidrule{2-12}						
& GDFQ & CNN&		26.86	&	42.35	&	54.15	&	55.55	&	63.35	&	61.62	&	69.52	&	61.58	&	70.07	\\ 						
& Qimera & CNN&		12.38	&	36.83	&	29.72	&	55.04	&	55.80	&	60.93	&	71.18	&	62.56	&	64.17	\\ 						
& AdaDFQ & CNN&		30.10	&	43.77	&	51.26	&	57.53	&	61.37	&	58.98	&	71.34	&	64.87	&	71.65	\\ 						
&PSAQ-ViT V1 & ViT &		61.19	&	54.11	&	43.83	&	65.52	&	57.24	&	61.17	&	71.74	&	66.58	&	73.87	\\ 						
& PSAQ-ViT V2 & ViT &		63.37	&	65.22	&	74.05	&	67.22	&	73.97	&	75.32	&	74.37	&	71.30	&	75.97	\\ \cmidrule{2-12}						
& \multirow{2}{*}{\aname (Ours)}  & \multirow{2}{*}{ViT} &		69.79	&	75.69	&	81.92	&	68.39	&	76.36	&	80.26	&	77.87	&	78.45	&	81.34	\\						
&&&		+6.42	&	+10.47	&	+7.87	&	+1.17	&	+2.39	&	+4.93	&	+3.50	&	+7.15	&	+5.38	\\ \midrule						
																									
\multirow{8}{*}{W4/A8}& Real-Data FT & - &		71.52	&	78.63	&	82.89	&	69.99	&	78.08	&	80.10	&	79.78	&	80.59	&	83.14	\\ \cmidrule{2-12}						
& GDFQ & CNN&		41.14	&	57.31	&	68.45	&	61.05	&	70.42	&	72.45	&	74.09	&	72.66	&	77.73	\\ 						
& Qimera & CNN&		34.22	&	60.08	&	63.22	&	61.90	&	70.10	&	72.38	&	73.93	&	72.22	&	76.35	\\ 						
& AdaDFQ & CNN&		41.27	&	57.56	&	68.69	&	62.79	&	70.90	&	71.61	&	74.97	&	73.88	&	78.83	\\ 						
&PSAQ-ViT V1 & ViT &		59.59	&	62.98	&	67.74	&	65.01	&	71.50	&	73.96	&	74.41	&	75.29	&	79.51	\\ 						
& PSAQ-ViT V2 & ViT &		66.78	&	72.54	&	80.13	&	66.03	&	77.01	&	79.06	&	77.40	&	77.25	&	80.59	\\ \cmidrule{2-12}						
& \multirow{2}{*}{\aname (Ours)}  & \multirow{2}{*}{ViT} &		68.15	&	75.02	&	82.39	&	68.16	&	77.29	&	80.48	&	78.09	&	79.39	&	82.55	\\						
&&&		+1.37	&	+2.49	&	+2.26	&	+2.13	&	+0.27	&	+1.41	&	+0.69	&	+2.14	&	+1.96	\\ \midrule						
																									
\multirow{8}{*}{W8/A8}& Real-Data FT & - &		74.76	&	77.85	&	81.22	&	71.40	&	77.72	&	78.80	&	79.30	&	79.47	&	83.00	\\ \cmidrule{2-12}						
& GDFQ & CNN&		47.92	&	67.25	&	75.71	&	68.82	&	75.16	&	76.40	&	76.29	&	74.43	&	79.33	\\ 						
& Qimera & CNN&		51.68	&	65.39	&	0.00	&	68.05	&	72.86	&	73.87	&	76.14	&	74.48	&	78.07	\\ 						
& AdaDFQ & CNN&		60.60	&	64.52	&	70.93	&	69.27	&	73.96	&	73.65	&	76.49	&	75.24	&	80.02	\\ 						
&PSAQ-ViT V1 & ViT &		71.67	&	70.14	&	68.77	&	71.25	&	75.58	&	75.81	&	76.29	&	77.37	&	81.70	\\ 						
& PSAQ-ViT V2 & ViT &		73.05	&	75.60	&	80.09	&	71.42	&	76.89	&	77.34	&	77.54	&	77.60	&	81.13	\\ \cmidrule{2-12}						
& \multirow{2}{*}{\aname (Ours)}  & \multirow{2}{*}{ViT} &		74.25	&	79.74	&	83.97	&	71.52	&	78.99	&	81.39	&	79.68	&	81.48	&	83.51	\\						
&&&		+1.20	&	+4.14	&	+3.87	&	+0.10	&	+2.10	&	+4.05	&	+2.13	&	+3.89	&	+1.80	\\ \bottomrule															

    \end{tabular}}
        \caption{Experimental results on ImageNet image classification dataset with LSQ~\cite{esser2019learned} activation quantization.}
    \label{tab:lsq_full}
\end{table*}

\end{document}